\documentclass[lettersize,journal]{IEEEtran}
\usepackage[caption=false,font=normalsize,labelfont=sf,textfont=sf]{subfig}

\usepackage{amsmath,amsfonts}
\usepackage{algorithmic}
\usepackage{algorithm}
\usepackage{array}
\usepackage[caption=false,font=normalsize,labelfont=sf,textfont=sf]{subfig}
\usepackage{textcomp}
\usepackage{stfloats}
\usepackage{url}
\usepackage{verbatim}
\usepackage{graphicx}
\usepackage{cite}
\usepackage{multirow,booktabs}
\usepackage{amsmath}
\usepackage{algorithm}
\usepackage{algorithmic}
\usepackage{color}
\usepackage[table,HTML]{xcolor}
\usepackage{enumitem}
\usepackage{pifont}
\usepackage[colorlinks, linkcolor=red, citecolor=blue]{hyperref}

\usepackage{ragged2e} 
\usepackage{threeparttable}
\usepackage{multirow} 

\begin{document}

\title{Leverage Cross-Attention for End-to-End Open-Vocabulary Panoptic Reconstruction\\}

\author{Xuan Yu, Yuxuan Xie, Yili Liu, Sitong Mao, Shunbo Zhou, Haojian Lu, Rong Xiong, Yiyi Liao, Yue Wang 
\thanks{Xuan Yu, Yuxuan Xie, Yili Liu, Haojian Lu, Rong Xiong, Yiyi Liao, and Yue Wang are with Zhejiang University, Hangzhou, Zhejiang, China. Yue Wang is the corresponding author {\tt\footnotesize wangyue@iipc.zju.edu.cn}}%
\thanks{$^{2} $Sitong Mao and Shunbo Zhou  with Huawei Cloud Computing Technologies Co., Ltd., Shenzhen, China.}}




\maketitle

\begin{abstract}
    Open-vocabulary panoptic reconstruction offers comprehensive scene understanding, enabling advances in embodied robotics and photorealistic simulation. In this paper, we propose PanopticRecon++, an end-to-end method that formulates panoptic reconstruction through a novel cross-attention perspective. This perspective models the relationship between 3D instances (as \textit{queries}) and the scene’s 3D embedding field (as \textit{keys}) through their attention map.
    Unlike existing methods that separate the optimization of queries and keys or overlook spatial proximity, PanopticRecon++ introduces \textit{learnable 3D Gaussians} as instance queries. This formulation injects 3D spatial priors to preserve proximity while maintaining end-to-end optimizability. Moreover, this query formulation facilitates the alignment of 2D open-vocabulary instance IDs across frames by leveraging optimal linear assignment with instance masks rendered from the queries. Additionally, we ensure semantic-instance segmentation consistency by fusing query-based instance segmentation probabilities with semantic probabilities in a novel panoptic head supervised by a panoptic loss.  During training, the number of instance query tokens dynamically adapts to match the number of objects. PanopticRecon++ shows competitive performance in terms of 3D and 2D segmentation and reconstruction performance on both simulation and real-world datasets, and demonstrates a user case as a robot simulator. Our project website is at: \href{https://yuxuan1206.github.io/panopticrecon_pp/}{https://yuxuan1206.github.io/panopticrecon\_pp/}
    
\end{abstract}


\section{Introduction} \label{sec:info}



\IEEEPARstart{P}{anoptic} reconstruction provides a comprehensive understanding of the environment with integrated 3D geometry, appearance, semantics, and instance information, which is highly valuable for embodied robotics in human robot interaction and photo-realistic simulation. In particular, open-vocabulary panoptic reconstruction allows for the perception of any object class in the open world, making the robot plug-and-play.

One intuitive but expensive method for open-vocabulary 3D panoptic reconstruction is to reconstruct and then segment 3D space by a 3D vision language model (VLM)\cite{xu2025pointllm, huang2023chat-3d, zhu20233d-vista, qi2025shapellm, huang2024chat-scene}, which requires a large amount of high-quality 3D annotated data.
In contrast, 2D annotated data is far more accessible, enabling 2D open-vocabulary panoptic segmentation to reach near-application levels. Therefore, some feature-lifting methods ~\cite{peng2023openscene, kerr2023lerf, dff, add1, add2} propose to lift the capabilities of 2D VLMs to a 3D representation. 2D VLM features~\cite{openseg, lseg, sam, clip} of images are distilled to a 3D visual language feature field by neural rendering \cite{mildenhall2021nerf} or back projection. While using the inner product of text prompts and distilled features enables flexible segmentation, its simple architecture results in imprecise masks.

\begin{figure*}[t]
    \centering
    \includegraphics[width=\linewidth]{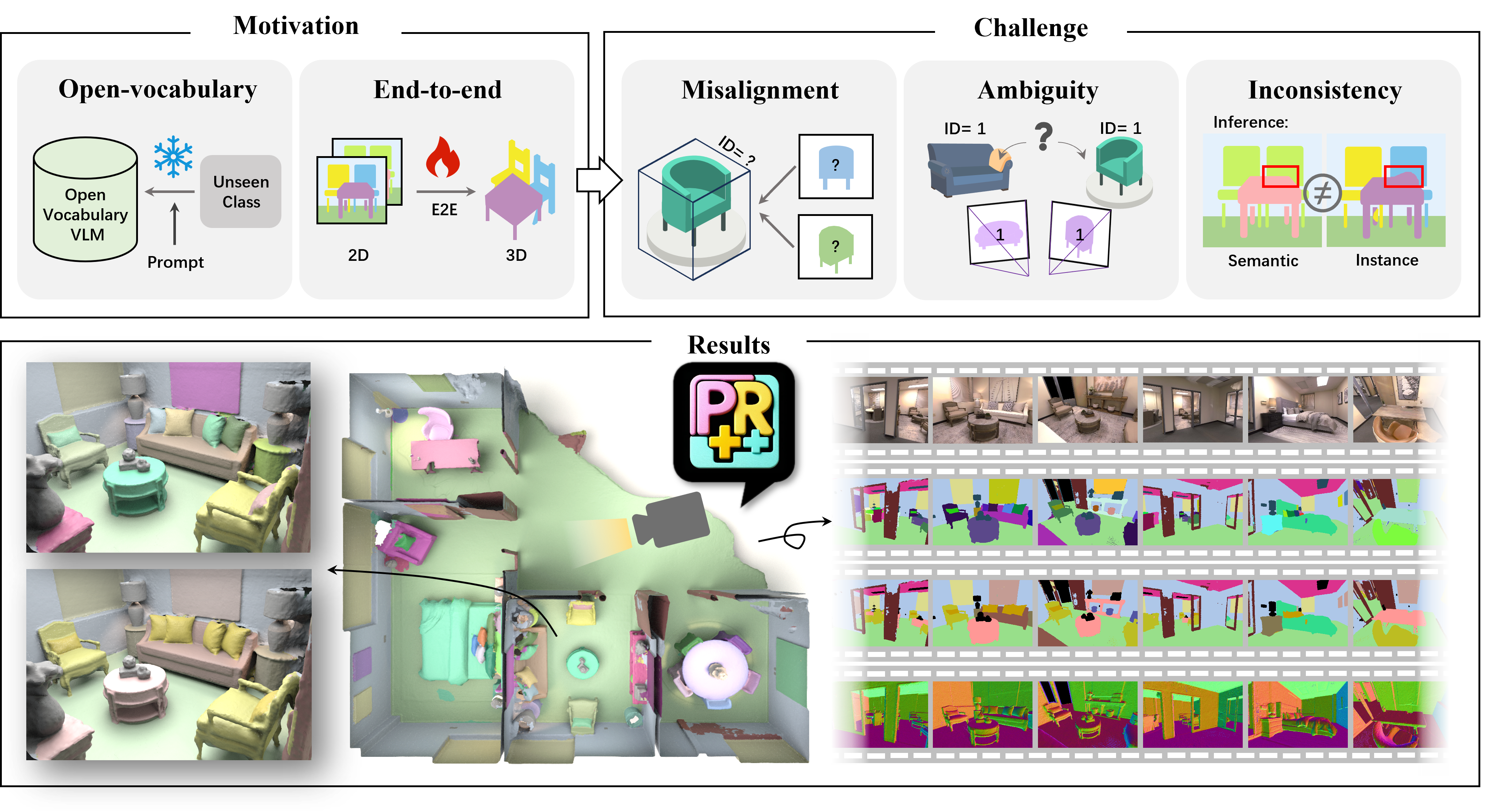}
    \caption{End-to-end open-vocabulary panoptic reconstruction by 2D foundation model faces three challenges:  1) Misalignment: 2D instance IDs across frames are not aligned. 2) Ambiguity: Due to the limited FoV, two objects that never co-occur in a single image can be the same or different instances. 3) Inconsistency: The semantic and instance segmentations obtained from two separated heads are inconsistent. We align 2D instance IDs by instance tokens linear assignment, eliminate the ambiguity of 3D instances by incorporating spatial prior, and output consistent semantic and instance masks by a parameter-free panoptic head, generating the geometric mesh with panoptic masking that allows for multi-branch novel-view synthesis.}
    \label{fig:challenge}
    \vspace{-0.3cm}
\end{figure*}

To improve the 3D segmentation performance, approaches \cite{takmaz2023openmask3d, lu2023ovir, nguyen2024open3dis, yin2024sai3d, Chen2024PVLFF, yu2024panopticrecon} shift to mask-lifting, which directly lift results (i.e., masks)  of the 2D VLM rather than features.  Leveraging high-quality 2D masks, the lifted 3D representation exhibits superior robustness and boundary accuracy. Despite the promising results, these mask-lifting methods face three major challenges as shown in Fig.~\ref{fig:challenge}: 1) Misalignment: 2D instance IDs across frames are not aligned in the whole data sequence. 2) Ambiguity: Due to the limited field of view (FoV), it is ambiguous to determine whether objects that never co-occur in a single image belong to the same instance. 3) Inconsistency: Existing methods usually use two separate heads to model semantic and instance labels, yielding inconsistent results.

We introduce a cross-attention~\cite{vaswani2017attention} perspective to analyze existing mask-lifting methods.
By encoding 3D instances as \textit{queries} and the embedding field of the scene as \textit{keys}, the final binary instance mask is derived from the \textit{attention map}. Following this way, there are two lines of works. The first line learns the queries and keys in two stages. The representative work, Contrastive Lift \cite{bhalgat2023contrastive} first learns the instance embedding field (\textit{keys}) using a contrastive loss defined by 2D instance masks; second, segment instance (\textit{queries}) by clustering the embeddings~\cite{mcinnes2017hdbscan}.
This approach introduces a proximity spatial prior to 3D instance via clustering for disambiguation. But it is not end-to-end, thus sensitive to parameter tuning to cluster a 3D instance. 
The second line learns the queries and keys at the same time. The representative work, Panoptic Lifting \cite{lifting}, aligns 2D instance IDs via linear assignments~\cite{kuhn1955hungarian} between 2D and 3D instances. Then the lifting problem becomes multi-class classification, with the classifier weights and instance embedding field being \textit{queries} and \textit{keys}, achieving end-to-end learning. However, due to the difficulty in incorporating 3D spatial priors in classifier weights, two 3D instances may have the same ID due to the ambiguity of instance.
Moreover, both lines employ separate semantics and instances branches without the supervision of unified panoptic segmentation, causing inconsistency between semantic and instance masks.


In this paper, we propose \textit{PanopticRecon++}, an open-vocabulary panoptic reconstruction method following the cross-attention perspective to address the limitations of previous works. Cross-attention provides a natural way to model the relationship between 3D instances and 3D embedding field of the scene by treating the former as \textit{query} and the latter as \textit{key}.
We introduce \textit{learnable 3D Gaussians} as instance queries, injecting 3D spatial priors to preserve proximity while maintaining end-to-end optimizability by jointly learning queries and keys.
Specifically, each query token is rendered to 2D and aligned with 2D instance IDs across frames through linear assignment~\cite{kuhn1955hungarian}. 
By incorporating segmentation features and spatial prior into the attention map, we disambiguate instance IDs caused by limited field of view.
We construct a parameter-free panoptic head by fusing the instance probability of query tokens and the semantic probability of semantic branch to define a panoptic loss, thereby ensuring semantic-instance consistency. Furthermore, to adaptively adjust the number of instance tokens, we present a dynamic token adjustment method that considers token relationships in 2D and 3D space to remove and add tokens. Finally, as shown in Fig.~\ref{fig:challenge}, PanopticRecon++ generates the geometric mesh with panoptic masking that allows for multi-branch novel-view synthesis, demonstrating superior geometry and segmentation accuracy compared with existing methods in experiments. We summarize our contributions as follows: 
\begin{itemize}
    \item We investigate open-vocabulary panoptic segmentation from the perspective of cross-attention, providing insights into the relationship between the instances and the scene. 
    
    \item We propose an end-to-end instance branch with 3D instance spatial prior by a set of adjustable 3D Gaussian-modulated instance tokens. 
    
    \item We present a parameter-free panoptic head to ensure consistency between semantic and instance labels.
    
    \item We evaluate the method for 3D/2D segmentation and mesh geometric quality on substantial scenes and demonstrate a user case as a robot simulator.

\end{itemize}

Our work is an extension version of a conference paper \cite{yu2024panopticrecon}, where we introduce PanopticRecon, a two-stage zero-shot panoptic reconstruction method. 
The most fundamental distinction between PanopticRecon++ and PanopticRecon lies in the shift to a fully end-to-end architecture. 
PanopticRecon leverages the 2D VLMs and the geometric priors provided by graph inference and shows competitive performance. 
In contrast, the key idea of PanopticRecon++ is to perform segmentation by introducing instance tokens modeled as 3D Gaussians. A cross-attention mechanism jointly leverages spatial priors and feature cues.
Specifically, we have several extensions: 1) We propose a perspective of cross-attention to unify the previous methods and delve into new insight for architecture design; 2) We present PanopticRecon++, which injects instance prior and build panoptic supervision, and keeps the whole architecture end-to-end differentiable. 3) We conduct substantial experiments for 2D/3D segmentation, reconstruction, and rendering tasks on simulated and real-world indoor datasets.

\section{Related Works}

\subsection{Close-vocabulary Segmentation Reconstruction}

Early studies, including Kimera \cite{rosinol2020kimera} and SemanticFusion \cite{mccormac2017semanticfusion}, adapt pre-trained 2D semantic segmentation models to process fundamental spatial representations (e.g., occupancy, SDF, or point clouds) for semantic 3D mapping, often relying on global optimization techniques such as bundle adjustment and Bayesian updates. Kimera further incorporates predicted 3D bounding boxes for 3D instance segmentation. Nonetheless, the performance of these methods is primarily limited by the model's capacity and the update fusion strategy. 
Recent studies leverage implicit neural representations \cite{semanticnerf,pnf,lifting,bhalgat2023contrastive,panopticnerf,nfatlas} or explicit Gaussian representations \cite{zhu2024sni-slam}, mapping 2D/3D semantic instance ground truth or predictions from 2D segmentation pre-trained models \cite{mask2former} to a unified 3D feature space. Semantic NeRF~\cite{semanticnerf} first explored encoding semantics into NeRF by adding a separate branch to predict semantic labels, fusing noisy 2D semantic segmentation into a consistent volumetric model. 
PNF~\cite{pnf} first proposed a method to obtain panoptic radiation fields from semantic images and the 3D bounding boxes of instances. To avoid the strong dependency on expensive 3D bounding boxes, most current methods choose to lift 2D labels to the 3D space \cite{wang2022dm, lifting, bhalgat2023contrastive} for 3D instance semantic segmentation. Contrastive Lift \cite{bhalgat2023contrastive} proposes learning segmentation features via contrastive learning and then obtaining 3D instance segmentation through post-processing clustering. DM-Nerf \cite{wang2022dm} and Panoptic Lifting \cite{lifting} employ optimal supervision association to end-to-end learn 3D instances from 2D instance masks. However, they are only supervised by 2D segmentation without any spatial prior, limited by the 2D field of view, resulting in an incomplete understanding of the 3D space. In order to incorporate 3D spatial priors, Instance-NeRF \cite{instancenerf} aligns 2D instance IDs by extracting discrete 3D masks from RGB density.
But the accuracy of extracting discrete 3D masks is susceptible to variations across different scenes, which can adversely affect the alignment of instance IDs. Despite the significant progress, these close-vocabulary methods remain confined to predefined closed-set category lists in specific datasets, performing poorly in open-world settings. This limitation hinders their ability to understand unseen object categories in complex and open-ended scenes.

\subsection{Open-vocabulary Segmentation}
\medskip \noindent\textbf{2D Segmentation.} 
Trained on massive image-text datasets, foundation models like CLIP \cite{clip} have achieved remarkable performance in aligning visual and textual representations. This lead to a surge of research exploring various zero-shot image tasks, such as object detection \cite{liu2023grounding} and image captioning \cite{mokady2021clipcap,li2022blip}, leveraging CLIP features. Building on this foundation,  LSeg \cite{lseg} and OpenSeg \cite{openseg} extend the foundation models to semantic image segmentation. Then, Segment Anything Model (SAM) \cite{sam} further advances to segment arbitrary objects in any image, driven by user-provided or automatic generation prompts. This highlights the model's ability to learn a generalizable notion of objects. More recently, Grounded SAM \cite{ren2024grounded} innovatively combines object detection and segmentation foundation models to enable zero-shot 2D semantic and instance segmentation by text prompts. Additionally, several methods~\cite{cheng2023tracking,SNNtrack,SNN} perform tracking on 2D images to achieve consistent multi-frame 2D segmentation.

\medskip \noindent\textbf{3D Segmentation.} 
Inspired by the 2D open-vocabulary segmentation foundation models, OpenScene \cite{peng2023openscene} proposes an open-vocabulary 3D semantic segmentation by distilling the CLIP feature onto 3D point clouds. 
These methods with a distillation of CLIP feature allow open-vocabulary querying with text prompts but cannot differentiate between instances of the same semantic class. To tackle this problem, several approaches \cite{takmaz2023openmask3d, huang2025openins3d, nguyen2024open3dis, yin2024sai3d, yan2024maskclustering, yu2024panopticrecon} leverage a combination of 3D priors and 2D generalizable features for open-vocabulary 3D instance segmentation. On the one hand, proposal-based top-down methods \cite{takmaz2023openmask3d, huang2025openins3d} utilize pre-trained Mask3D \cite{schult2023mask3d} to generate 3D masks as proposals for 2D instance ID alignment. On the other hand, bottom-up methods \cite{yu2024panopticrecon, yan2024maskclustering, nguyen2024open3dis, yin2024sai3d} propose over-segmenting the mesh based on normals, and then merging the over-segmented surface according to their projection relationships with 2D instance masks to achieve 3D open-vocabulary instance segmentation. Both methods are not end-to-end, resulting in a potential accumulation of errors. 

\subsection{Open-vocabulary Segmentation Reconstruction}
Thanks to volume rendering-based reconstruction methods \cite{mildenhall2021nerf,wang2021neus,muller2022instant} supervised by images, it is possible to achieve open-vocabulary segmentation and reconstruction simultaneously leveraging  2D foundation models.
LERF \cite{kerr2023lerf} and DFF \cite{dff} distill CLIP feature to a feature field to integrate language into 3D space, enabling flexible queries at any 3D point. However, these feature-lifting methods \cite{kerr2023lerf,dff,Chen2024PVLFF} still suffer from the inability to distinguish instance objects, rough segmentation boundaries, and high memory usage.
When integrated with segmentation images from 2D foundation models, mask-lifting methods \cite{yu2024panopticrecon,kim2024garfield,gaussian_grouping} can achieve more precise segmentation and reconstruction.
GARField \cite{kim2024garfield} lifts hierarchical 2D instance segmentation to 3D by contrastive learning. It optimizes the instance neural field such that pixels belonging to the same 2D mask are pulled closer and otherwise pushed apart, mitigating the problem of misaligned 2D instance mask IDs. Multi-scale segmentation reconstructions can be achieved, but this method requires post-processing to cluster features and may have difficulty in segmenting all instances well with a single parameter set. 
Gaussian Grouping \cite{gaussian_grouping} directly supervises 3D models with object-level 2D instance IDs from Grounded SAM \cite{ren2024grounded} to ensure 3D object-level instance reconstruction. However, it faces the challenge of misaligned 2D instance IDs. Gaussian Grouping uses a 2D tracker to align instance IDs, but tracking yields unstable performance.
To align 2D instance IDs more precisely, we introduce geometric priors to understand 3D scenes comprehensively. In our previous work \cite{yu2024panopticrecon}, we combined geometric priors with 3D graph inference to align 2D masks, which was achieved through a two-stage approach. In this work, we leverage the cross-attention between neural field and Gaussian-modulated tokens, combine the segmentation feature with the geometric prior, and propose an end-to-end framework for open-vocabulary panoptic reconstruction.


\begin{figure*}[t]
    \centering
    \includegraphics[width=\linewidth]{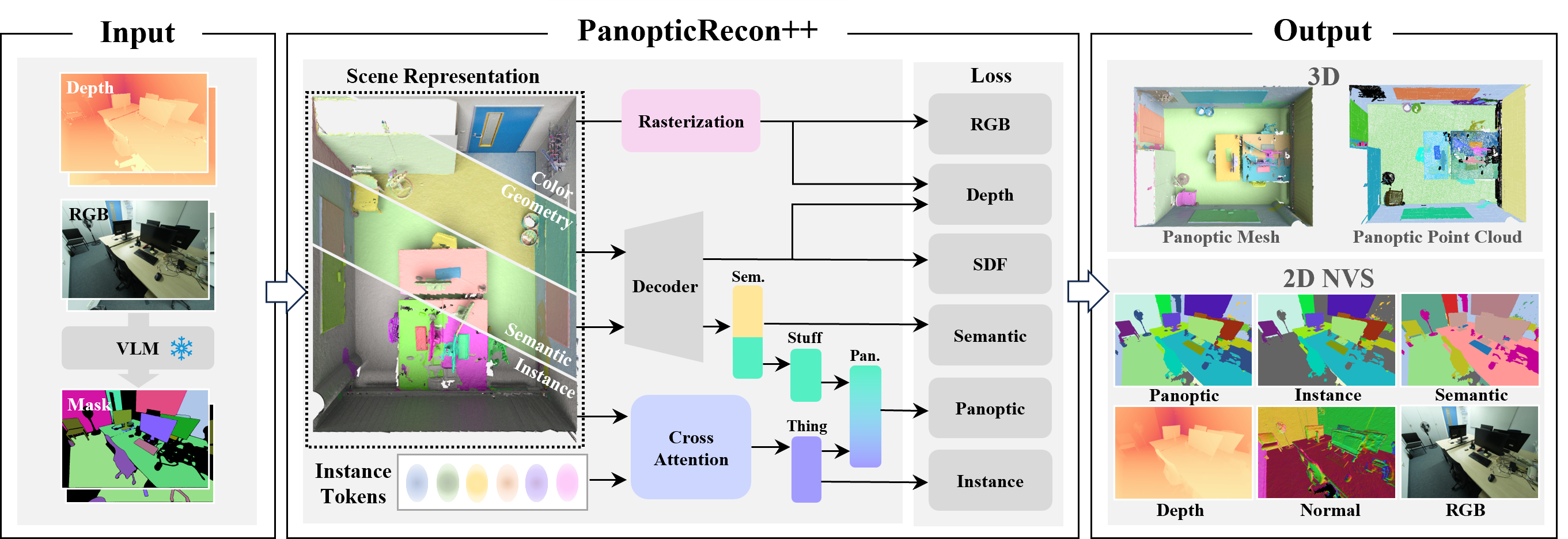}
    \caption{The input to PanopticRecon++ is posed RGB-D and segmentation images generated by Grounded SAM \cite{ren2024grounded}. The field representation comprises appearance, SDF, semantics, and instances. Appearance leverages 3DGS \cite{kerbl20233dgs}, and we use three hierarchical hashed encoding models \cite{muller2022instant} for SDF, semantics, and instances. The radiance field is supervised by RGB loss and depth loss. The geometry field is supervised by depth loss and SDF loss. The probabilities of the output stuff class from the semantic field and the instance probabilities computed from the instance field and instance tokens through cross-attention are concatenated under Bayes' rule to form the panoptic (Pan.) probability. The segmentation images generated by Grounded SAM directly supervise semantic, instance, and panoptic probabilities. Finally, PanopticRecon++ outputs high-quality panoptic mesh, point cloud, and multi-branch novel-view synthesis.}
    \label{fig:overview}
    \vspace{-0.3cm}
\end{figure*} 

\section{A Cross-attention Perspective} 
In this section, we first detail the instance representation from a cross-attention perspective, and then present our system overview.

Cross-attention aggregates features based on the similarity between two types of tokens. 
In segmentation models employing cross-attention, each instance is represented by a query token $Q_i$, while each point in 3D space is represented by a key token $K_j$. The corresponding value $V_j$ represents the semantic feature of that point. The attention map $\mathcal{A}_{ij}$, indicating the similarity between $Q_i$ and $K_j$, is typically computed using a softmax function over all points. The instance class feature $F_i$ is then aggregated as follows:
\begin{equation}
F_i = \sum_j \mathcal{A}_{ij} V_j = \sum_j \frac{\exp{Q_i^T K_j}}{\sum_j \exp{Q_i^T K_j}} V_j 
\label{ca}
\end{equation}
where $\mathcal{A}_{ij}$ are applied for instance mask prediction, while $F_i$ is applied for class prediction.

\subsection{Analysis of 3D Fields in Mask-lifting} 
From a cross-attention perspective, we analyze existing 3D panoptic reconstruction methods that leverage 2D VLM masks, categorizing them into two approaches: mask-guided feature field learning, and mask-guided ID field learning.

\medskip \noindent\textbf{Feature Field Learning.} Methods of this type, such as \cite{bhalgat2023contrastive, Chen2024PVLFF, kim2024garfield} employ a two-stage segmentation pipeline. The first stage uses a mask-guided contrastive loss to learn a feature field. The second stage clusters this 3D feature field into instances. From a cross-attention perspective, $Q$ encodes both coordinates $p_q$ and features $f_q$ of cluster centroids, and $K$ encodes the per-point coordinates $p_k$ and features $f_k$ within the feature field. $\mathcal{A}_{ij}$ is defined as follows:
\begin{equation}
    \mathcal{A}_{ij} = d(Q_i(\bar{f}_q, p_q),K_j(\bar{f}_k, \bar{p}_k))
    \label{ffl}
\end{equation}
where $d$ measures a combination of spatial distance between centroids and 3D points coordinates as well as the similarity between centroids and 3D points features. The specific form depends on the employed clustering method. 
This indicates that both VLM masks (represented by features $f_q$ and $f_k$) and the 3D spatial prior (represented by coordinates $f_q$ and $f_k$) are considered by these methods. Note that $\bar{f}_q$, $\bar{f}_k$ and $\bar{p}_k$ denote features learned using a proxy loss (i.e., the contrastive loss in the first stage), while $p_q$ are learned in the second stage with other parameters frozen. Thus, these methods are not end-to-end and require post-processing that is sensitive to parameter tuning.

\medskip \noindent\textbf{ID Field Learning.} This line of methods \cite{wang2022dm,lifting} uses a softmax classifier to predict per-ID instance masks, supervised by a cross-entropy loss. 2D instances are aligned with the ID field by linear assignment. 
In this way, the weights of the softmax classifier serve as $Q$, and $K$ is per-point ID embedding in the ID field. $\mathcal{A}_{ij}$ is defined as standard softmax-based attention:
\begin{equation}
    \mathcal{A}_{ij} = \frac{\exp{Q_i(f_q)^T K_j(p_k)}}{\sum_j \exp{Q_i(f_q)^T K_j(p_k)}}
    \label{ifl}
\end{equation}
As the softmax weights vary only with instance ID, $Q_i$ is independent of spatial coordinates. For $K_j$, only spatial coordinates are used to learn the ID embedding. Therefore, the attention map in these methods does not consider both spatial prior and VLM masks in the query and key. A key advantage is that the model is trained in a single stage, with direct backpropagation from the instance segmentation mask loss.

\medskip \noindent\textbf{Insight to Bridge the Gap.} Based on the preceding analysis, the insight about an ideal instance branch is to combine the advantages of both lines of methods: Encode both spatial prior and mask induced features in both query and key as (\ref{ffl}), while aligning the ID for one-stage learning as (\ref{ifl}). $\mathcal{A}_{ij}$ is then formulated as:
\begin{equation}
    \mathcal{A}_{ij} = \frac{\exp{Q_i(f_q,p_q)^T K_j(f_k,p_k)}}{\sum_j \exp{Q_i(f_q,p_q)^T K_j(f_k,p_k)}}
    \label{ours}
\end{equation} 
Details regarding the instance branch design that realizes this formulation are provided in Sec.~\ref{section:token}.

An intuitive choice for the value $V$ is semantic field, which encodes the class of each 3D point. However, given that we only have 2D mask observations, which cover part of instance, thus 3D points belonging to one instance may not be fully supervised. This can lead to a single instance being misclassified as different classes across different images. As there is no need for generalization in reconstruction tasks, the design of value can be outside the convention of cross-attention. As shown in Eq.~(\ref{ca}), the purpose of value aggregation is to derive $F_i$, which encodes instance-level class. We propose to directly learn $F_i$ for each query instance and assign all points belonging to that instance the unique class predicted by $F_i$~\cite{mask2former, schult2023mask3d}, disregarding $V_j$ for each point, as shown in Fig. \ref{fig:prior}.

\subsection{System Overview} 
Inspired by cross-attention, we introduce an instance branch to our system, enabling end-to-end learning for panoptic segmentation. As illustrated in Fig.~\ref{fig:overview}, PanopticRecon++ comprises a multi-branch scene representation and instance tokens.

\medskip \noindent\textbf{Input with 2D VLM.} \label{sec:VLM}
Given an RGBD sequence, we leverage class text prompts to guide VLM (Grounded SAM \cite{ren2024grounded}) to generate 2D instance and semantic segmentations, which provide the supervision for the panoptic reconstruction task. 
When pixels are assigned multiple instance labels, we select the label with the highest confidence score. We group the instances with the same class into semantic masks and divide all semantic classes into \textit{thing} and \textit{stuff} categories. Note that the per-frame instances are initially unassociated. They are aligned with the instance masks rendered by the instance branch during training.

\medskip \noindent\textbf{Scene Representation.} As illustrated in Fig.~\ref{fig:overview}, PanopticRecon++ employs 3D Gaussian-modulated instance tokens and represents the scene using four fields: appearance, geometry, instance and semantics. We model appearance using Gaussian Splatting \cite{kerbl20233dgs} and represent per-point geometry, semantics, and instance embeddings with three neural volumes. Appearance is rendered using rasterization, while depth and semantics are rendered via neural volume rendering. Instances are rendered using cross-attention-based neural rendering. These rendered outputs are self-supervised by observations to train the fields.

\medskip \noindent\textbf{Panoptic Segmentation Head.} Thanks to the differentiable instance branch, PanopticRecon++ allows for end-to-end learning guided by panoptic segmentation through the combination of semantic and instance branches.
We propose a parameter-free panoptic head following Bayes rule \cite{bayes1958bayes} for fusion based on the probabilities of \textit{stuff}, \textit{thing}, and instance.
The \textit{stuff} and \textit{thing} class probabilities are directly extracted from the semantic segmentation probabilities output by the neural semantic field. Instance probabilities are derived from the binary mask probabilities computed by the attention mechanism between instance tokens and the instance field. During training, the semantic, instance, and panoptic head are supervised using 2D masks from the VLM. At inference time, as illustrated in Fig. \ref{Fig_panoptic}, instance segmentation is obtained by excluding the \textit{stuff} category from the panoptic segmentation. Each instance token corresponds to a semantic label. By replacing the instance IDs of \textit{thing} category objects in the panoptic segmentation with their corresponding semantic labels, semantic segmentation is derived.


\section{Instance Branch Design} \label{section:token}

\begin{figure}
    \centering
    \includegraphics[width=\linewidth]{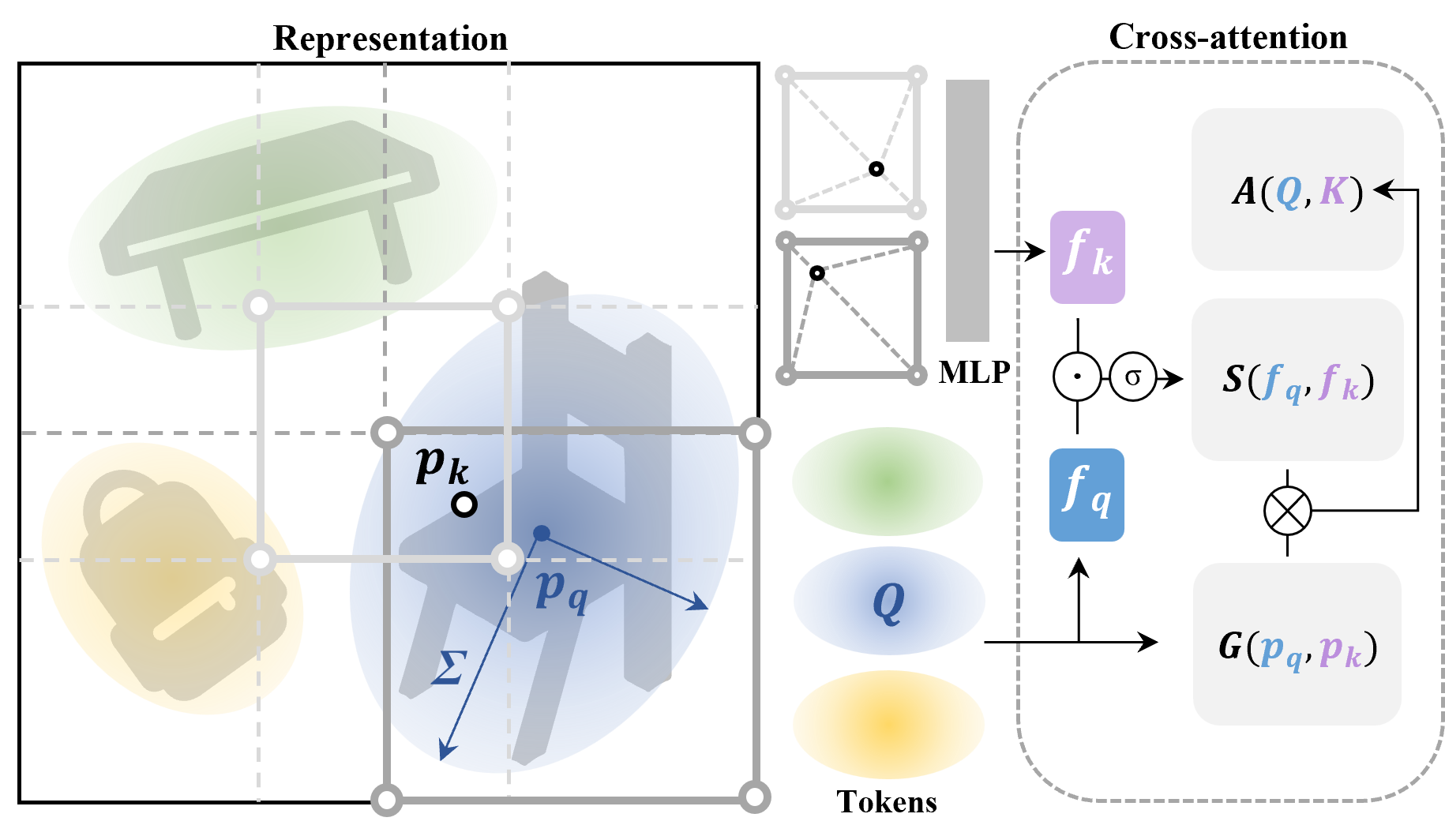}
    \caption{Instance branch design by cross-attention between 3D Gaussian-modulated query token and spatial hashing encoded scene fields. }
    \label{fig:token}
    \vspace{-0.3cm}
\end{figure}

This section details the specific architecture of the instance branch. As illustrated in Fig.~\ref{fig:token}, query tokens consist of instance features $f_q$ and spatial coordinates $p_q$, modulated by 3D Gaussian distributions with learnable attributes. Key tokens consist of spatial coordinates $p_k$, and instance features $f_k$ encoded in a voxel grid indexed by spatial hashing. 3D instances masks and classes are predicted by cross-attention between the query and key tokens.

\subsection{3D Gaussian-modulated Query Token} \label{sec:token} 

Given the diverse shapes of open-vocabulary objects, fitting fine-grained object shapes as priors using methods designed for specific instance types \cite{engelmann2016shapeprior} is challenging. To address this issue, we introduce a learnable ellipsoid shape prior for instance tokens,  implemented as a 3D Gaussian-modulated token (Fig.~\ref{fig:token}). Each query token encodes a spatial 3D Gaussian distribution with center $p_q$ and covariance matrix $\Sigma_q\in \mathbb{R}^{3 \times 3}$, and also incorporates an instance feature $f_q$. Augmenting the query defined in Eq.~(\ref{ours}), the 3D Gaussian-modulated instance token is parameterized as $Q_i(f_q, p_q| \Sigma)$. In our experiments, we find the 3D Gaussian-modulated token provides a rational instance spatial prior and simplifies the complex task of establishing shape prior. 


\medskip \noindent\textbf{Differentiability.}
To ensure the positive semi-definite covariance matrix \(\Sigma\) \cite{kerbl20233dgs} differentiable, we decompose it as:
\begin{equation}
    \Sigma = R \Lambda \Lambda^T R^T
\end{equation}
where the rotation matrix $R \in \mathbb{R}^{3 \times 3}$ is compactly parameterized by a quaternion $q \in \mathbb{R}^{4}$, and the diagonal matrix $\Lambda \in \mathbb{R}^{3 \times 3}$ is concisely expressed as a 3D vector $\lambda \in \mathbb{R}^{3}$. This decomposition ensures that the covariance matrix remains positive semi-definite during optimization.


\medskip \noindent\textbf{Initialization.}
To efficiently initialize instance tokens and mitigate optimization challenges, we generate initial tokens near the surfaces of 3D instances. Specifically, we randomly sample pixels from the 2D instance masks and project them into 3D space using the depth map to obtain the initial token positions $p_q$.

\begin{figure}
    \centering
    \includegraphics[width=\linewidth,keepaspectratio]{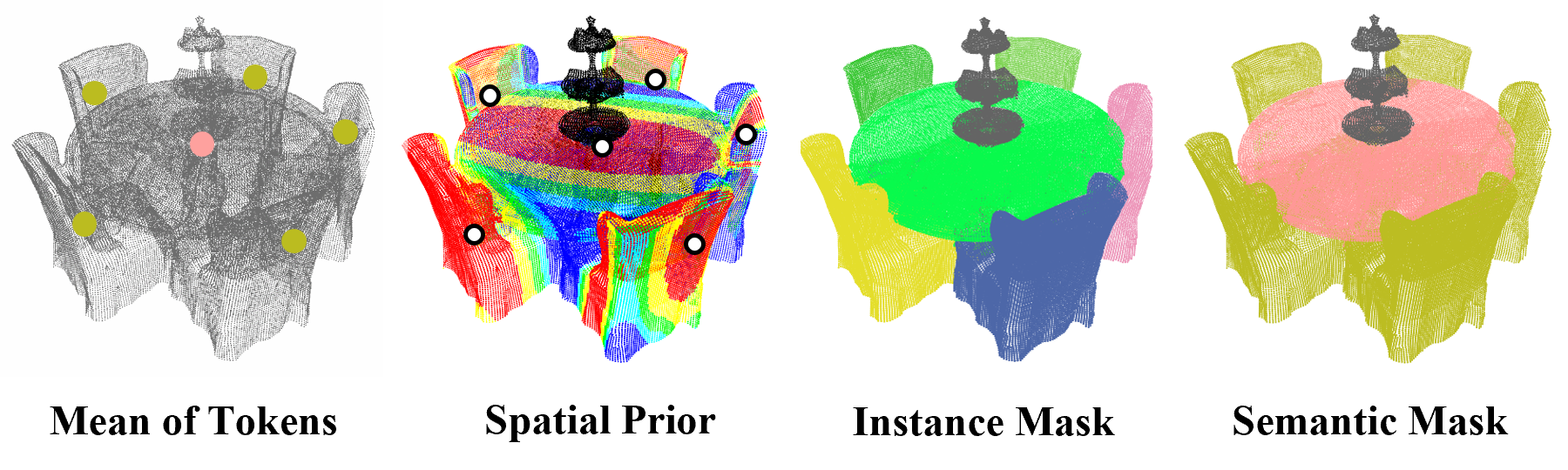}\\
    \caption{\label{Fig_prob} A visualization of the contribution of various attributes of instance tokens to segmentation. Instance classes are visualized by semantic label colors on the mean points of instance tokens. The intensity of the spatial prior is visualized using a color gradient, with red indicating the highest intensity, decreasing outwards.}
    \vspace{-0.3cm}
    \label{fig:prior}
\end{figure} 

\subsection{Cross-Attention Design}
As the spatial prior is encoded as a Gaussian distribution, the inner product-based similarity measure in Eq.~(\ref{ours}) cannot be directly applied. We model the feature similarity $\mathcal{S}$, modulated by the spatial similarity $\mathcal{G}$ between the instance tokens and each scene point, resulting in the following attention map $\mathcal{A}$:
\begin{equation}
\mathcal{A}_{ij} = \frac{\exp{\mathcal{S}(f_q,f_k) \mathcal{G}(p_q,p_k)}}{\sum_j \exp{\mathcal{S}(f_q,f_k) \mathcal{G}(p_q,p_k)}}
\label{gsours}
\end{equation}
The feature similarity $S$ is defined as:
\begin{equation}
    \mathcal{S}(f_q,f_k) = \sigma(f_q^T f_k)
\end{equation}
where $\sigma$ denotes the sigmoid function, used to normalize the similarity. The spatial prior $\mathcal{G}$ is defined as:
\begin{equation}
    \mathcal{G}(p_q,p_k) = \frac{P(p_k|p_q,\Sigma)}{P(p_q|p_q,\Sigma)}
    \label{eq:wight}
\end{equation}
where $P(\cdot|\cdot)$ denotes the Gaussian probability density with a specified mean and covariance, representing the spatial proximity to an instance. To normalize across different instance tokens, we divide $P(p_k|p_q,\Sigma)$ by its peak $P(p_q|p_q,\Sigma)$, yielding a normalized spatial prior $\mathcal{G}(p_q,p_k) \in [0,1]$, where higher values indicate stronger spatial affinity. This normalization ensures the prior reflects relative spatial compatibility rather than absolute density, making it suitable for modulating feature attention weights over the 3D scene.

The visualization of $\mathcal{G}$ is shown in Fig. \ref{fig:prior}. The Gaussian modulation follows the assumption that points of an instance are concentrated in a localized region around its centroid $p_q$. By modulating  $\mathcal{S}$ with this Gaussian prior, the resulting attention map approximates the clustering behavior of Eq.~ (\ref{ffl}), while maintaining differentiability. Compared to Eq.~(\ref{ifl}), this spatial prior aided attention avoids isolated objects never observed together in a single image being incorrectly grouped as a single instance.

\subsection{Dynamic Token Adjustment}
Given the number of instances varies across scenes, the number of instance tokens must be automatically adjusted during training. To this end, we propose a pruning and splitting mechanism for instance tokens, coupled with the corresponding decision criteria.


\medskip \noindent\textbf{Pruning.} 
During optimization, We prune redundant tokens, which fall into two categories: useless and duplicate.

Useless tokens are those that are never assigned or assigned only infrequently. As our image instance masks are derived from imperfect open-vocabulary segmentation results rather than ground-truth annotations, infrequent assignment of a token likely indicates an erroneous mask. Therefore, we prune these tokens to reduce computational cost.

Duplicated tokens refer to multiple tokens are assigned with the same instance. Due to the limited FoV of 2D images, some tokens may match only portions of one instance during early optimization. As optimization progresses, one token converges to represent the complete instance mask, leaving the others redundant. To identify these duplicate tokens, we define a mask coverage metric, 3D Intersection over Mask (IoM). Given a token $Q_i$ with 3D mask $M_i$, if there exists a token $Q_j$ whose 3D mask $M_j$ intersects with $M_i$, IoM is defined as:
\begin{equation}
    \text{IoM}_{ij} = \frac{M_i \cap M_j}{M_i}
\end{equation}
which evaluates the 3D overlap between each pair of instance tokens.
We observe that duplicate tokens often exhibit an enclosing pattern, where a larger 3D instance mask encompasses multiple smaller 3D instance masks. So we keep only the token with the largest mask. 


        

\medskip \noindent\textbf{Split.}
Ideally, as training converges, tokens form a one-to-one correspondence with 3D instances, yielding optimal assignments for 2D masks. When the initial token set is sparse, a single token may be the lowest-cost match for multiple nearby objects, causing persistent assignment conflicts. We address this by monitoring assignments and duplicating tokens on demand: Given masks $M_i, M_j$ from the same image and a token $Q_i$, if the assignment cost function $\mathcal{C}(Q_i,M_i)=\min_k \mathcal{C}(Q_k,M_i)$ and $\mathcal{C}(Q_i,M_j)=\min_k \mathcal{C}(Q_k,M_j)$, yet the assignment assigns $Q_i$ to $M_i$ and forces $M_j$ to some $Q_\ell\neq Q_i$ with $\mathcal{C}(Q_\ell,M_j)>\mathcal{C}(Q_i,M_j)$ over several iterations, we clone $Q_i$ as $Q_i'$ and append it to the token set, alleviating competition and enabling both $M_i$ and $M_j$ to obtain optimal matches.

\section{End-to-end Panoptic Reconstruction} \label{section:system}
This section focuses on the neural image synthesis and design of loss function, for geometric reconstruction, panoptic segmentation, and appearance. As Fig. \ref{fig:overview} shows, PanopticRecon++ represents a scene using four fields: appearance, geometry, semantics, and instances. 
Due to hierarchical hashed encoding (HHE) \cite{muller2022instant} feature volume excels at capturing geometric details while offering efficient parameter storage and point querying, geometry and semantic fields are constructed by two separate HHE volumes, followed by a small multilayer perceptron (MLP) decoder. Instance field is also constructed using HHE feature volumes, but the small MLP is replaced by the cross-attention between the instance field and instance tokens presented in Sec.~\ref{section:token}.
Appearance field is populated with 3D Gaussians \cite{kerbl20233dgs} to efficiently fit scene complex texture. 
Furthermore, a parameter-free panoptic head, that connects semantic and instance fields, is introduced to predict panoptic labels directly, aligning semantic and instance masks from two branches.

\subsection{Neural Image Synthesis}
The feature volumes for geometry $G$, semantics $S$, and instances $I$, are denoted as $\Psi_{v}$, $v\in\{G,S,I\}$. $\Psi_{v}$ are supervised at the ray level, thus we introduce the volume rendering to generate the prediction of the pixel corresponding to the ray. Based on the camera pose, we define the origin and direction of a ray $(o,d)$ passing through a pixel $x$ in an RGB, depth, semantic or instance segmentation image. Along the ray, we sample $N$ successive 3D spatial points $\{p_j\}$ as $p_j = o + \rho_j d$, where $\rho_j$ is the distance. Based on the neural fields, each point has the prediction $f(p_j)$, which can be SDF $s$, normal $n$, depth $d$, semantic probability $l_{S}$, instance probability $l_{I}$, and panoptic probability $l_{P}$. 

\medskip \noindent\textbf{Geometry Branch.}
We have the volume rendering as follows:
\begin{equation}
    u_f(x) = \sum_{i=j}^{N}{T_{j} \alpha_{j} f(p_{j})}
    \label{vr}
\end{equation}
where $T_{j}=\prod_{m=1}^{j-1}(1-\alpha_{m})$, $\alpha_{m}$ is the discrete opacity value defined under the S-density function assumption~\cite{wang2021neus} as:
\begin{equation}
    \alpha_{j} = \max\left(\frac{ \sigma(s_j)-\sigma(s_{j+1}) }{ \sigma(s_j) }, 0\right) 
    \label{weight}
\end{equation}
where $s$ is SDF of the point, $\sigma(s)$ is Sigmoid function $\sigma(s)=(1+e^{-\xi s})^{-1}$ with a temperature coefficient $\xi$.

Given a 3D point $p_j$, the geometry feature volume maps $p_j$ to a feature vector $\Psi_{G}(p_j)$. A small MLP $h_G$ is applied to yield the SDF value $s \in \mathbb{R}$:
\begin{equation}
    s(p_j) = h_{G}(\Psi_{G}(p_j))
\end{equation}
Its normal $n\in \mathbb{R}^3$  is derived as:
\begin{equation}
    n(p_j) = \frac{\partial s}{\partial p_j}
\end{equation}

\medskip \noindent\textbf{Semantic \& Instance Branches.}
We maintain two separate neural feature volumes for semantic and instance segmentation to ensure the fitting capacity of each model. 

Given $C$ classes, the semantic label probability distribution of $p_j$ denoted as $l_S(p_j) \in \mathbb{R}^C$ is generated by querying the semantic volume feature $\Psi_{S}(p_j)$, which is passed through a small MLP $h_S$ as:
\begin{equation}
    l_S(p_j)= h_{S}(\Psi_{S}(p_j))
\end{equation}

Given $N$ instance queries, unlike the semantic branch, the instance ID probability at $p_j$ denoted as $l_I(p_j) \in \mathbb{R}^N$, which is $N$ binary predictions for $N$ queries, thus not normalized. It is computed via cross-attention (\ref{ours}) between $N$ query tokens $Q(f_q, p_q)$ and the instance volume feature $\Psi_{I}(p_j)$: 
\begin{equation}
    l_I(p_j)= \left[\mathcal{A}_{1j} \ldots \mathcal{A}_{ij} \ldots \mathcal{A}_{Nj} \right]
\end{equation}


\medskip \noindent\textbf{Appearance Branch.}
Since the appearance branch employs Gaussian Splatting, we spatially align its representation with geometry volume representations through depth supervision. The depth supervision also reduces the number of floating Gaussians in free space caused by insufficient observations, improving the quality of rendered images from a novel view.

The appearance of the scene is explicitly represented with a set of 3D Gaussians \cite{kerbl20233dgs}. Each Gaussian is defined by a Gaussian distribution. Given a transformation matrix $W$ and an intrinsic matrix $K$, $p_q$ and $\Sigma$ of a 3D Gaussian $G$ can be transformed to camera coordinate corresponding to $W$ and then projected to 2D coordinate:
\begin{equation}
    p_q' = [x',y',z']^T = KW[p_q, 1]^T
\end{equation}
\begin{equation}
    \Sigma' = J W \Sigma W^T J^T 
\end{equation}
where $J$ is the Jacobian of the affine approximation for the $W$. Rendering color \cite{kerbl20233dgs} and depth \cite{cheng2024gaussianpro} of a pixel $x$ can be obtained by $\alpha$-blending, where the color is represented by spherical harmonics. 

\subsection{Panoptic Image Synthesis}

\begin{figure}[t]
    \centering
    \includegraphics[width=\linewidth,keepaspectratio]{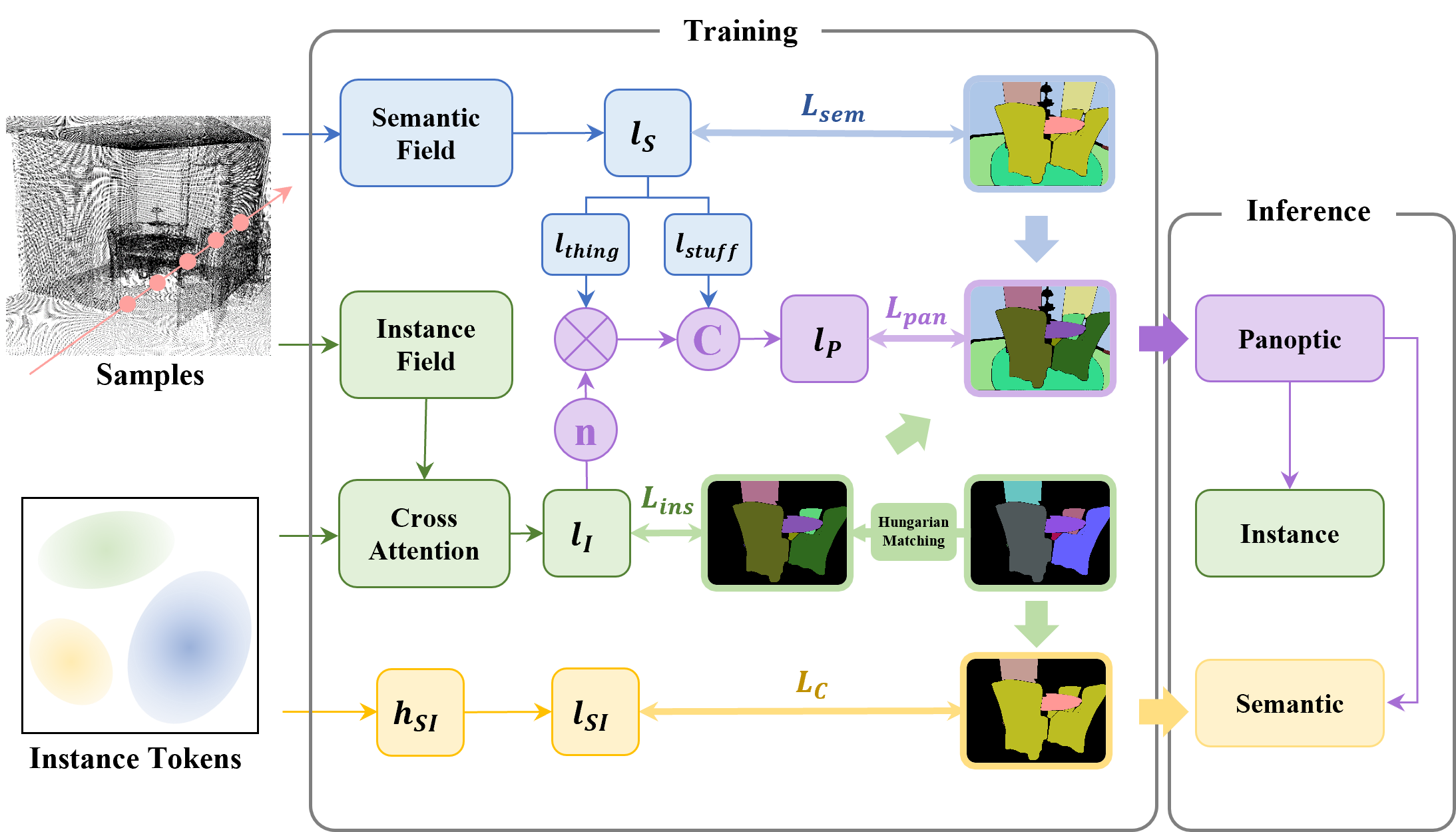}\\
    \caption{\label{Fig_panoptic} The architecture of the panoptic segmentation head in both training and inference stages. During training, $L_{sem}$ and $L_{ins}$ maintain the semantic branch (\fbox{\color[HTML]{B4C7E7}\rule{0.2cm}{0.2cm}{}}) and instance branch (\fbox{\color[HTML]{BCDDA7}\rule{0.2cm}{0.2cm}{}}), respectively. The semantic classes of the instance tokens are supervised by $L_{C}$. The parameter-free panoptic head, derived from the fusion of the semantic and instance branches, is trained using $L_{pan}$, enabling direct prediction of panoptic probability. During inference, semantic and instance segmentation results are directly derived from the panoptic segmentation output of the parameter-free panoptic head.} 
    \vspace{-0.3cm}
\end{figure}

Traditional 3D panoptic segmentation methods typically employ a two-stage post-processing approach, querying instance labels within semantic segmentation foreground masks \cite{yu2024panopticrecon, lifting, Chen2024PVLFF}. Inspired by recent 2D panoptic segmentation architecture that predicts panoptic labels directly \cite{cheng2021maskformer, xiong2019upsnet}, we propose a 3D panoptic segmentation head built upon semantic and instance branches, enabling end-to-end learning and ensuring semantic and instance consistency.

As shown in Fig. \ref{Fig_panoptic}, the probability $l_S$ from the semantic branch is split to $l_{stuff}$ and $l_{thing}$ respectively, corresponding to the probabilities of \textit{stuff} and \textit{thing} classes. 
To achieve end-to-end fusion of semantic and instance segmentation probabilities, we follow Bayes rule \cite{bayes1958bayes} rather than simply concatenating the results. Given a point $p_j$, its panoptic label is:
\begin{equation}
    l_P(p_j) = \left[l_{stuff}(p_j), \sum l_{thing}(p_j) \cdot \frac{l_I(p_j)}{\sum l_I(p_j)}\right]
    \label{panoptic}
\end{equation}
where $\sum l_{thing}$ means the probability that the semantic class of $p_j$ is thing, $\frac{l_I(p_j)}{\sum l_I(p_j)}$ means the probability that $p_j$ belongs to any instances conditioned on the semantic class of $p_j$ is thing. Then we show that the panoptic label defined in (\ref{panoptic}) is a valid probability distribution:
\begin{align}
    &\sum l_{stuff} +  \sum \sum l_{thing} \cdot \frac{l_I}{\sum l_I} \\
    =& \sum l_{stuff} + \sum l_{thing} \sum \cdot \frac{l_I}{\sum l_I}\\
    =& \sum l_{stuff} + \sum l_{thing} = 1
\end{align}
Since the distribution is normalized to $1$, and all entries are larger than $0$, we complete the validness of the label.


\medskip \noindent\textbf{Instance Class Prediction.}
Finally, to train the instance class feature $F_i$, we also add an instance class prediction branch as:
\begin{equation}
    l_{SI}(Q_i) = h_{SI}(F_i)
    \label{eq:instance class}
\end{equation}
where $h_{SI}$ indicates a small MLP for class prediction. As shown in Fig.~\ref{Fig_panoptic}, when querying for semantic class of a given point, its panoptic label directly tells the probability for stuff classes. To derive the probability for thing classes of the query point, we group its instance probability by instance classes.




\subsection{End-to-end Learning}
With all point-level predictions, we can generate pixel-level predictions by Eq.~\ref{vr}. We set up segmentation loss, geometric loss, and appearance loss to supervise the panoptic reconstruction. In a nutshell, the total loss $L$ is: 
\begin{equation}
    L = L_{depth}+L_{sdf}+L_{eik}+L_{ins}+L_{sem}+L_{pan}+L_c+L_{rgb}
\end{equation}

\medskip \noindent\textbf{Geometric Loss.}
The geometric loss consists of three parts:
\begin{equation}
    L_{sdf}(p_i) =\begin{cases}
                \vert s(p_i) - b(p_i)\vert  & |b|\leq \tau\\
                \max(0, e^{-\beta s(p_i)}-1, s(p_i) - b(p_i)) & o.w.
                \end{cases}
\end{equation}
where $\tau$ is a threshold to truncate SDF, $b(p_i)=D(x)-\rho_i$ is the distance between the $p_i$ and the observed depth of pixel $x$ along the ray as an approximated SDF. In addition, eikonal loss is used as regularization:
\begin{equation}
    L_{eik}(p_i) = \|1-|n(p_i)|\|^2
\end{equation}

Depth is also supervised at the pixel level for depth generated by volume rendering:
\begin{equation}
    L_{depth}(x) = \vert D(x) - u_d(x)\vert \label{eq:depth_loss}
\end{equation}

\medskip \noindent\textbf{Instance Assignment.}
To address the challenge of misaligned 2D instance IDs across frames, we adopt a common approach \cite{wang2022dm, lifting} of aligning predicted instance labels with ground-truth labels using the Hungarian Algorithm \cite{kuhn1955hungarian} during 2D mask supervision, enabling end-to-end 2D instance ID supervision.
Considering the unknown number of instances in a scene, we transform the pixel-wise multi-class classification problem into a binary classification problem for each token. 

Specifically, given an instance segmentation image, we can get $L$ 2D instance binary masks, denoted as $M^I=\{M^I_0, M^I_1, ..., M^I_L\}$. Corresponding to the same view, our instance branch predicts $N$ 2D instance binary masks ($N$ is the number of tokens), denoted as $\bar{M}^I=\{\bar{M}^I_0, \bar{M}^I_1, ..., \bar{M}^I_N\}$. We construct a cost matrix $C \in \mathbb{R}^{L \times N}$ ($L \leq N$) as:
\begin{equation}
C = L_{dice} + L_{bce}
\end{equation}
where $L_{dice}$ is a metric used to quantify the similarity between the predicted instance binary mask probability $\bar{M}^I$ and the target instance binary matrix $M^I$:
\begin{equation}
L_{dice}(\bar{M}^I, M^I) = 1 - \frac{2 \bar{M}^I M^I + \epsilon}{\bar{M}^I + M^I + \epsilon}
\end{equation} 
where \(\epsilon\) is a smoothing term introduced to enhance numerical stability, prevent division by zero, and mitigate the impact of small sample sizes. We set $\epsilon=1$ in our experiments. $L_{bce}$ measures the discrepancy between the predicted probability distribution and the ground truth distribution:
\begin{equation}
L_{bce}(\bar{M}^I, M^I) = -(M^I \log{\bar{M}^I} + (1 - M^I) \log(1 - \bar{M}^I))
\end{equation} 
The cost matrix is then fed into the Hungarian algorithm \cite{kuhn1955hungarian} to find the optimal assignment between the elements of $M^I$ and $\bar{M}^I$, resulting in the matched target 2D instance binary mask ${M^I}'$.


\medskip \noindent\textbf{Instance Loss.}
After aligning each pair of ${M^I}'$ and $\bar{M}^I$, $L_{ins}$ at the pixel $x$ is defined as:
\begin{equation}
L_{ins}(x) = L_{dice}(\bar{M}^I(x), {M^I}'(x)) + L_{bce}(\bar{M}^I(x), {{M^I}'(x)})
\label{eq:instance_loss}
\end{equation}
$L_{ins}$ supervises the binary masks of instance tokens for efficient learning of instance field and instance tokens. 
In addition, for instance class feature, we define $L_c$ as with the instance class prediction $l_{SI}$ (\ref{eq:instance class}) and $L$ 2D instance class masks $M_S$:
\begin{equation}
L_{c}(l_{SI},M^S) = -(M^S \log{(l_{SI}} + (1 - M^S) \log(1 - l_{SI})
\end{equation} 


\medskip \noindent\textbf{Semantic Loss.}
Each ray from $x$ yields a semantic probability distribution which is supervised by cross-entropy of $u_{l_S}$ to semantic image labels $M^s$:
\begin{equation}
    L_{sem}(x) = - \sum {M^S(x) \log u_{l_S}(x)}
\end{equation}
$L_{sem}$ maintains the semantic field, providing foreground and background probabilities as priors for the fusion of probabilities to build the panoptic head.

\medskip \noindent\textbf{Panoptic Loss.}
Similar to semantic supervision, the panoptic loss is a cross-entropy of $u_{l_P}$ to panoptic image labels $M^P$:
\begin{equation}
    L_{pan}(x) = - \sum {M^P(x) \log u_{l_P}(x)}
\end{equation}
$L_{pan}$ supervises the panoptic head, refining semantic and instance branches and balancing their contributions, ensuring consistent semantic and instance segmentation derived from panoptic segmentation during inference.

\medskip \noindent\textbf{Appearance Loss.}
The appearance loss $L_{rgb}$ consists of the image reconstruction loss $L1$ and $L_{SSIM}$ in 3DGS \cite{kerbl20233dgs} and depth loss $L_d$ same as Eq. \ref{eq:depth_loss} for aligning the geometry and appearance brunches:
\begin{equation}
    L_{rgb} = (1-\lambda)L1 + \lambda_1 L_{SSIM} + \lambda_2 L_{d} 
    \label{eq:rgb_loss}
\end{equation}
where $\lambda_1=0.2$ same with 3DGS and $\lambda_2=0.1$.


\subsection{Implementation}

Finally, we introduce the implementation of the whole system. In the 2D instance segmentation stage, we provide Grounded SAM with a comprehensive list of descriptive phrases or keywords corresponding to potential objects or stuff classes within each experimental setting. Each keyword is associated with a custom semantic label and a foreground/background class label. Multiple descriptive terms can be linked to a single semantic label to enhance detection accuracy. Furthermore, we employ a maximum detection probability confidence filtering mechanism to address over-segmentation issues. Specifically, for pixels that are assigned to multiple segments, the final label is determined based on the grounding probability of the most confident prediction.

During the training process, to ensure the stability of the model training, we only use $L_{depth}+L_{sdf}+L_{eik}+L_{rgb}$ at the first epoch. Once the geometry field is established, the SDF, through Eq. \ref{weight}, can provide more accurate weights for sampling points, which enhances the efficiency of training for the semantic and instance branches. Then all losses except $L_{pan}$ are activated, and the four branches are jointly optimized. After two epochs, the direct supervision of the panoptic loss $L_{pan}$ is introduced. Since the masks produced by Grounded SAM tend to shrink slightly from the true object boundaries, resulting in label gaps between adjacent masks, we remove $L_{dice}$ of $L_{ins}$ after the instance ID assignment becomes relatively stable, leaving $L_{bce}$ to mitigate over-reliance on boundaries and facilitate the completion of label gaps based on surrounding feature similarity.

\begin{table*}[t]
    \centering
    \setlength{\tabcolsep}{4pt}
    \caption{Panoptic Segmentation quality using different methods on Replica dataset}
    \centering
    \vspace{-0.3cm}
    \resizebox{\linewidth}{!}{
    \begin{tabular}{lccc|cc|cc|cc|cc}
        \toprule
            \multirow{2}*{\textbf{Method}} & \multicolumn{7}{c|}{2D-Scene} & \multicolumn{4}{c}{3D}\\
            \cmidrule(r){2-8} \cmidrule(r){9-12}
             & $\text{PQ}$$\uparrow$ & SQ$\uparrow$ & RQ$\uparrow$ 
             & mIoU$\uparrow$ & mAcc$\uparrow$ 
             & mCov$\uparrow$ & mW-Cov$\uparrow$ 
             & mIoU$\uparrow$ & mAcc$\uparrow$ 
             & mCov$\uparrow$ & mW-Cov$\uparrow$ \\
            
        \midrule
            Kimera
            & - & - & - 
            & - & - 
            & - & - 
            & 61.85 & 77.05 
            & - & - \\ 
            Panoptic NeRF
            & 62.11 & 69.22 & 89.69 
            & 60.54 & 75.80 
            & 62.46 & 74.25 
            & 49.22 & 71.78 
            & \textbf{69.21} & \textbf{94.40} \\ 
        \cmidrule(r){2-12}
            Panoptic Lifting
            & 52.95 & 60.40 & 71.18 
            & \cellcolor{orange!30}79.89 & 87.44 
            & 40.16 & 47.35 
            & 70.57 & 83.21 
            & 28.66 & 34.54 \\ 
            Contrastive Lift
            & 51.78 & 66.40 & 65.99 
            & 79.59 & \cellcolor{orange!30}87.96 
            & 24.75 & 34.23 
            & \cellcolor{orange!30}70.67 & \cellcolor{orange!30}83.29 
            & 18.49 & 26.56 \\ 
            PVLFF 
            & 48.74 & 67.75 & 65.93 
            & 56.14 & 66.03 
            & 52.32 & 67.44 
            & 41.77 & 51.54 
            & 45.33 & 57.12 \\ 
            PanopticRecon 
            & \cellcolor{orange!30} 70.75 & \cellcolor{orange!30}77.59 & \cellcolor{orange!30}84.04 
            & 75.37 & 81.78 
            & \cellcolor{orange!30}69.49 & \cellcolor{orange!30}86.81 
            & 65.88 & 77.52 
            & \cellcolor{orange!30}58.70 & \cellcolor{orange!30}72.52 \\ 
            \textbf{PanopticRecon++} 
            & \cellcolor{red!30} \textbf{80.04} & \cellcolor{red!30}\textbf{84.74} & \cellcolor{red!30}\textbf{94.69} 
            & \cellcolor{red!30}\textbf{82.82} & \cellcolor{red!30}\textbf{90.46} 
            & \cellcolor{red!30}\textbf{76.34} & \cellcolor{red!30}\textbf{88.32} 
            & \cellcolor{red!30}\textbf{70.69} & \cellcolor{red!30}\textbf{84.44} 
            & \cellcolor{red!30}\textbf{66.96} & \cellcolor{red!30}\textbf{75.20} \\ 
        \bottomrule
    \end{tabular}
    }
    \label{tab:seg_replica}
\end{table*}
\begin{table*}[t]
    \centering
    \setlength{\tabcolsep}{4pt}
    \caption{Panoptic Segmentation quality using different methods on ScanNet-V2 dataset}
    \centering
    \begin{threeparttable}
    \vspace{-0.3cm}
    \resizebox{\linewidth}{!}{
    \begin{tabular}{lccc|cc|cc|cc|cc}
        \toprule
            \multirow{2}*{\textbf{Method}} & \multicolumn{7}{c|}{2D-Scene} & \multicolumn{4}{c}{3D}\\
            \cmidrule(r){2-8} \cmidrule(r){9-12}
             & $\text{PQ}$$\uparrow$ & SQ$\uparrow$ & RQ$\uparrow$ 
             & mIoU$\uparrow$ & mAcc$\uparrow$ 
             & mCov$\uparrow$ & mW-Cov$\uparrow$ 
             & mIoU$\uparrow$ & mAcc$\uparrow$ 
             & mCov$\uparrow$ & mW-Cov$\uparrow$ \\

        \midrule
            Kimera 
            & - & - & - 
            & - & - 
            & - & - 
            & 70.67 & 84.52
            & - & - \\ 
            Panoptic NeRF
            & 58.19 & 64.44 & 78.77 
            & 59.12 & 70.86 
            & 63.46 & 72.02 
            & 55.56 & 70.88 
            & 72.36 & \textbf{88.14} \\ 
        \cmidrule(r){2-12}
            Panoptic Lifting
            & 57.89 & 61.96 & \cellcolor{orange!30}85.31 
            & 67.91 & 78.59 
            & 45.88 & 59.93
            & \cellcolor{orange!30}67.21 & \cellcolor{orange!30}82.96 
            & 52.96 & 55.22 \\ 
            Contrastive Lift
            & 37.35 & 41.91 & 57.60 
            & 64.77 & 75.80 
            & 13.21 & 23.26 
            & 66.36 & 82.89 
            & 13.89 & 18.39 \\ 
            PVLFF 
            & 30.11 & 42.79 & 44.43 
            & 55.41 & 63.96 
            & 45.75 & 48.41 
            & 46.60 & 57.89 
            & 48.44 & 47.09 \\ 
            PanopticRecon 
            & \cellcolor{orange!30}63.70 & \cellcolor{orange!30}64.81 & 81.17 
            & \cellcolor{orange!30}68.62 & \cellcolor{orange!30}80.87 
            & \cellcolor{orange!30}66.58 & \cellcolor{orange!30}77.84 
            & 65.22 & 81.74 
            & \cellcolor{orange!30}75.32 & \cellcolor{orange!30}73.68 \\ 
            \textbf{PanopticRecon++} 
            & \cellcolor{red!30}\textbf{75.55} & \cellcolor{red!30}\textbf{76.18} & \cellcolor{red!30}\textbf{99.19} 
            & \cellcolor{red!30}\textbf{76.56} & \cellcolor{red!30}\textbf{86.12} 
            & \cellcolor{red!30}\textbf{72.45} & \cellcolor{red!30}\textbf{81.70} 
            & \cellcolor{red!30}\textbf{74.11} & \cellcolor{red!30}\textbf{85.64} 
            & \cellcolor{red!30}\textbf{88.59} & \cellcolor{red!30}\textbf{82.60} \\ 

        \bottomrule
    \end{tabular}
    }
    \label{tab:seg_scannet}
    \end{threeparttable}
    \vspace{-0.3cm}
\end{table*}

\begin{figure*}[t]
    \centering
    \includegraphics[width=\linewidth,keepaspectratio]{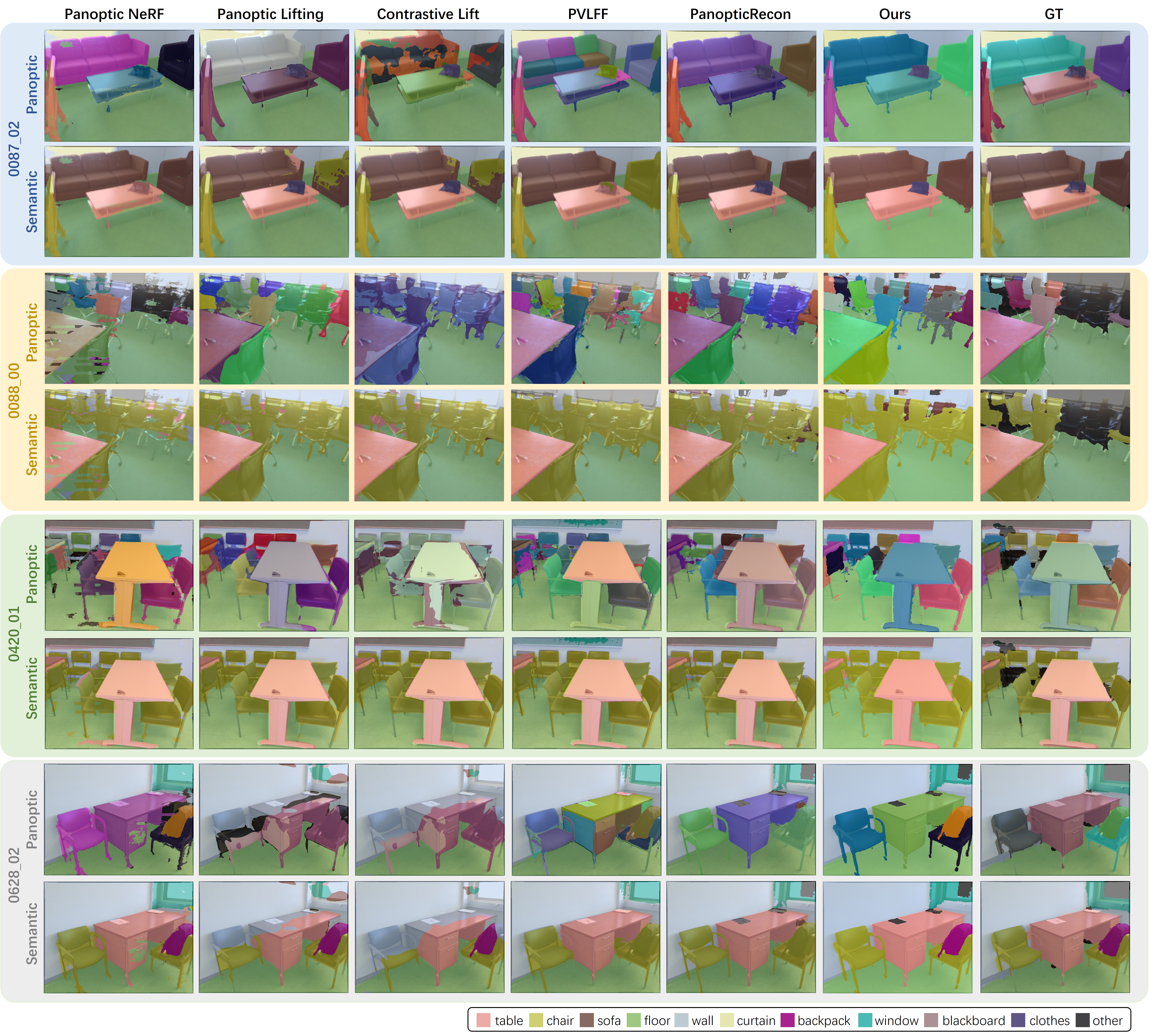}\\
    \vspace{-0.3cm}
    \caption{Comparison of the quality of semantic segmentation and panoptic segmentation of different methods on ScanNet.}
    \vspace{-0.3cm}
    \label{fig:seg_scannet}
\end{figure*}

\begin{table*}[t]
    \centering
    \setlength{\tabcolsep}{4pt}
    \vspace{-0.1cm}
    \caption{Panoptic Segmentation quality using different methods on ScanNet++ dataset}
    \centering
    \begin{threeparttable}
    \vspace{-0.3cm}
    \resizebox{\linewidth}{!}{
    \begin{tabular}{lccc|cc|cc|cc|cc}
        \toprule
            \multirow{2}*{\textbf{Method}} & \multicolumn{7}{c|}{2D-Scene} & \multicolumn{4}{c}{3D}\\
            \cmidrule(r){2-8} \cmidrule(r){9-12}
             & $\text{PQ}$$\uparrow$ & SQ$\uparrow$ & RQ$\uparrow$ 
             & mIoU$\uparrow$ & mAcc$\uparrow$ 
             & mCov$\uparrow$ & mW-Cov$\uparrow$ 
             & mIoU$\uparrow$ & mAcc$\uparrow$ 
             & mCov$\uparrow$ & mW-Cov$\uparrow$ \\

        \midrule
            Kimera 
            & - & - & - 
            & - & - 
            & - & - 
            & 71.27 & 85.19 
            & - & - \\ 
            Panoptic NeRF
            & 46.13 & 54.16 & 71.69 
            & 50.13 & 71.14 
            & 42.80 & 56.60 
            & 47.04 & 70.56 
            & 69.54 & \textbf{84.72} \\ 
        \cmidrule(r){2-12}
            Panoptic Lifting
            & \cellcolor{orange!30}71.15 & \cellcolor{orange!30}77.48 & \cellcolor{orange!30}88.14 
            \cellcolor{orange!30}& \cellcolor{orange!30}81.34 & \cellcolor{orange!30}89.67 
            & 56.17 & 68.51 
            & 73.01 & 85.76 
            & 65.20 & 66.38 \\ 
            Contrastive Lift
            & 47.58 & 57.23 & 65.81 
            & 81.09 & 89.30 
            & 27.39 & 36.51 
            & 70.98 & 85.20 
            & 32.13 & 33.94 \\ 
            PVLFF 
            & 52.24 & 66.86 & 65.56 
            & 62.53 & 70.31 
            & \cellcolor{orange!30}67.95 & \cellcolor{orange!30}75.47 
            & 53.27 & 63.49 
            & \cellcolor{orange!30}70.89 & \cellcolor{orange!30}69.56 \\ 
            PanopticRecon 
            & 68.29 & 77.01 & 85.00 
            & 77.75 & 87.08 
            & 51.34 & 62.79 
            & \cellcolor{orange!30}74.44 & \cellcolor{orange!30}86.80 
            & 61.47 & 61.36 \\ 
            \textbf{PanopticRecon++} 
            & \cellcolor{red!30}\textbf{77.28} & \cellcolor{red!30}\textbf{84.05} & \cellcolor{red!30}\textbf{91.09}  
            & \cellcolor{red!30}\textbf{81.47} & \cellcolor{red!30}\textbf{90.53} 
            & \cellcolor{red!30}\textbf{75.39} & \cellcolor{red!30}\textbf{78.91} 
            & \cellcolor{red!30}\textbf{75.95} & \cellcolor{red!30}\textbf{87.93} 
            & \cellcolor{red!30}\textbf{89.60} & \cellcolor{red!30}\textbf{80.04} \\ 
        \bottomrule
    \end{tabular}
    }
    \label{tab:seg_scannet++}
    \end{threeparttable}
    \vspace{-0.3cm}
\end{table*}

\begin{figure*}[t]
    \centering
    \includegraphics[width=\linewidth,keepaspectratio]{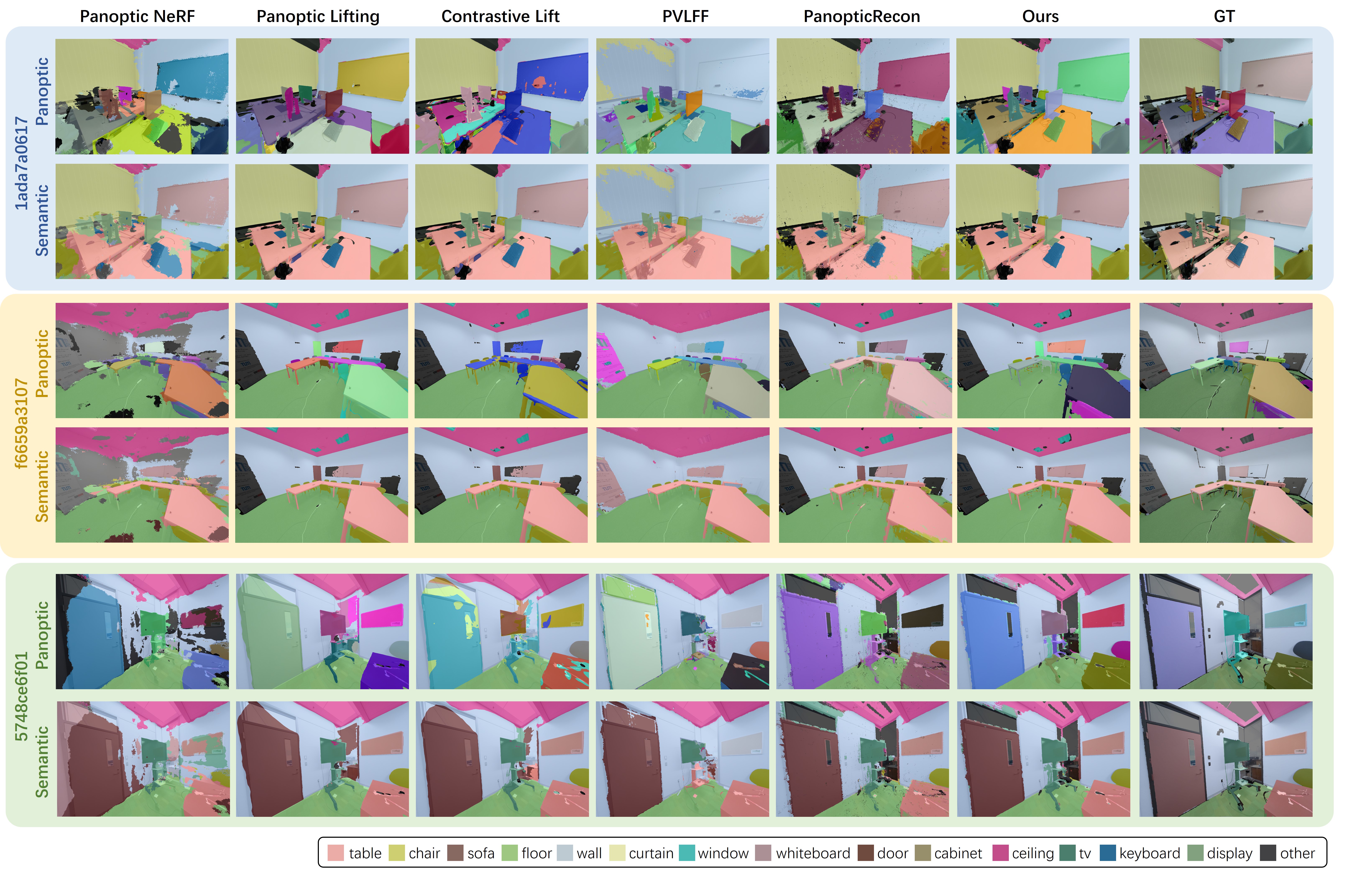}\\
    \vspace{-0.3cm}
    \caption{Comparison of the quality of semantic segmentation and panoptic segmentation of different methods on ScanNet++.}
    \vspace{-0.3cm}
    \label{fig:seg_scannet++}
\end{figure*}

\section{Experimental Results}
We first conduct comparative experiments on 2D/3D semantic, instance, and panoptic segmentation. Subsequently, we presented an comparison of reconstruction. Finally, we validated our design choices through ablation studies.

\subsection{Datasets}  \label{section:dataset}
We evaluate our approach on three indoor datasets and one outdoor dataset: one simulated dataset (Replica~\cite{straub2019replica}) and three real-world datasets (Scannet-V2~\cite{dai2017scannet} Scannet++~\cite{yeshwanth2023scannet++}, and KITTI-360~\cite{liao2022kitti}). 

\medskip \noindent\textbf{Replica.} 
The Replica dataset \cite{straub2019replica} contains 18 highly realistic 3D indoor scenes. We use a multi-room scene, "Apartment\_2", for our experiments. Since Replica doesn't provide camera poses, we manually specify keyframes and interpolate trajectories to generate RGB-D images and semantic instance segmentation ground truth. We use the dataset's high-precision mesh for reconstruction evaluation.

\medskip \noindent\textbf{Scannet-V2.}
We use the ScanNet-V2 dataset \cite{dai2017scannet}, an RGB-D sequence dataset of indoor scenes. It provides 3D camera poses, surface reconstruction meshes, and instance-level semantic annotations, making it suitable for evaluating reconstruction and segmentation. We use 4 scenes, “0087\_02”, "0088\_00", "0420\_01" and "0628\_02", from ScanNet-V2 for our experiments. The dataset features mesh reconstructed by BundleFusion \cite{dai2017bundlefusion} and semi-manually annotated instances and semantic labels. Compared to simulated data, ScanNet-V2 is affected by various real-world factors, such as blurring, exposure, noise, and manual errors, posing challenges for segmentation and reconstruction tasks. For evaluation, we adjust the list of semantic and instance labels for some of the ambiguous definitions and ensure that all methods followed the same list of labels during the evaluation. 

\medskip \noindent\textbf{Scannet++.} 
ScanNet++ dataset \cite{yeshwanth2023scannet++}, which contains high-resolution 3D scans and registered RGB images, has high-precision geometry and object-level semantic labels. We use the RGB images and depth maps to evaluate PanopticRecon++. We select “1ada7a0617”, "5748ce6f01", and "f6659a3107" scenes for our experiments. For evaluation, we use rendered semantic/instance images and meshes to assess segmentation and reconstruction performance. ScanNet++ also enables a new benchmark for novel view synthesis evaluation.

\medskip \noindent\textbf{KITTI-360.} 
To evaluate the generalization capability of our method on larger-scale scenes, we further conduct experiments on the outdoor KITTI-360 dataset \cite{liao2022kitti}. Specifically, we select a 200-meter segment from sequence 00 for evaluation, utilizing depth observations from a Velodyne HDL-64 LiDAR and a single perspective camera. Detailed experimental results and analysis are provided in the Appendix.

\subsection{Baseline}
We select several baselines for segmentation in both 2D and 3D space. First, Kimera \cite{rosinol2020kimera}, a learning-free approach for panoptic reconstruction, requires 2D semantic segmentation and 3D bounding boxes as input. Panoptic NeRF (PN) \cite{panopticnerf}, like Kimera, which also requires 3D object bounding boxes. The assumption of available 3D bounding boxes eliminates the instance association for both PN and Kimera, thus they are employed as a reference for the comparison.

Then Panoptic Lifting (PL) \cite{lifting}, Contrastive Lift (CL) \cite{bhalgat2023contrastive}, PVLFF \cite{Chen2024PVLFF}, and our previous work, PanopticRecon (PR) \cite{yu2024panopticrecon} are employed with the same problem setting to PanopticRecon++. They are representative works for panoptic reconstruction by lifting 2D observations, making them a fair comparison to PanopticRecon++ (PR++). Both CL and PVLFF are two-stage methods based on feature clustering post-processing. PR, our previous work, also adopts a two-stage reconstruction pipeline. PL is a one-stage method but has no instance prior consideration. 

\subsection{Evaluation Metrics}

\medskip \noindent\textbf{Panoptic Segmentation.} 
Following previous work \cite{lifting}, we first flatten and concatenate all panoptic segmentation images within a scene and employ the Hungarian algorithm to find an optimal matching between predicted and ground truth instances that maximizes the sum of IoU overall matched pairs. After establishing a correspondence between predicted panoptic masks and ground truth panoptic masks, we retain only the matched masks with an IoU greater than 0.5. This approach not only evaluates the segmentation performance on individual frames but also considers the inter-frame instance relationships. The recognition quality (RQ) evaluates the overlap between each pair of matched labels, and the semantic quality (SQ) assesses the correspondence between predicted and ground truth labels:
\begin{equation}
    RQ = \frac{|TP|}{|TP| + \frac{1}{2}|FP| + \frac{1}{2}|FN|}
\end{equation}
\begin{equation}
    SQ = \frac{\sum_{(p,g) \in TP} IoU(p, g)}{|TP|}
\end{equation}
The panoptic quality (PQ) is a comprehensive metric used to evaluate panoptic segmentation performance by considering both SQ and instance RQ:
\begin{equation}
    PQ = SQ \cdot RQ
\end{equation}
where $TP$ denotes the number of true positives, $TN$ denotes the number of true negatives, $FP$ denotes the number of false positives, and $FN$ denotes the number of false negatives. 

\medskip \noindent\textbf{Semantic Segmentation.} 
We evaluate the 2D/3D semantic segmentation with $mIoU$ and $mAcc$:
\begin{equation}
    IoU_i = \frac{TP_i}{TP_i + FP_i + FN_i}
\end{equation}
\begin{equation}
    mIoU = \frac{1}{C} \sum_{i=1}^{C} IoU_i
\end{equation}
\begin{equation}
    mAcc = \frac{1}{K} \sum_{i=1}^K \frac{TP_i + TN_i}{TP_i + FP_i + TN_i + FN_i}
\end{equation}
The regions in the scene where the ground truth semantic labels are \textit{'unlabeled'} or whose semantics are difficult to describe in the ground truth are set as unlabeled. In 2D semantic segmentation, $K$ represents the number of pixels to be evaluated, while in 3D semantic segmentation, it represents the number of point clouds to be evaluated. 

\medskip \noindent\textbf{Instance Segmentation.} 
For 2D instance segmentation, we obtain matched pairs between predicted and ground truth instance masks following the panoptic evaluation protocol. For 3D instance segmentation, we directly employ the Hungarian algorithm to associate instance-segmented point clouds with ground truth instance point clouds. Then we evaluate the 2D/3D instance segmentation with $mCov$ and $mW\text{-}Cov$:
\begin{equation}
    mCov = \frac{1}{O} \sum_{i=1}^{O} IoU_i
\end{equation}
\begin{equation}
    mW\text{-}Cov = \frac{\sum_{i=1}^{O} w_i \cdot IoU_i}{\sum_{i=1}^{O} w_i}
\end{equation}
where $O$ is the number of instances. $w_i$ is the ratio of the area (2D) or volume (3D) of the $i$-th ground truth instance to the total area or volume, thereby $mW\text{-}Cov$ emphasizing the segmentation performance on larger instances.

\medskip \noindent\textbf{Reconstruction.} 
We down-sample the vertices of the reconstructed mesh to a uniform number of points. Subsequently, we compute the average distance between reconstructed and ground truth points using the nearest neighbor search (K-means \cite{ahmed2020kmeans}) with meters as the unit. We employ a suite of metrics including Accuracy ($Acc.$), Completeness ($Com.$), Chamfer-L1 Distance ($C\text{-}L1$), Precision ($Pre.$), Recall ($Re.$), and F1-score ($F1$). $Acc.$, $Com.$, and $C\text{-}L1$ are all expressed in units of $cm$. The percentage thresholds for both $Pre.$, $Re.$ and $F1$ are set to $5 cm$. These metrics assess the reconstructed mesh's overall correctness, completeness, and consistency.

\medskip \noindent\textbf{Rendering.} 
We evaluate the color and depth of the rendered images separately. $PSNR$, $SSIM$, and $LPIPS$ are employed to comprehensively assess the quality of the rendered RGB images~\cite{zhang2018psnr}. Absolute Relative Error ($AbsRel$ \cite{eigen2014absrel}) is utilized to reflect the overall discrepancy between the rendered depth map and the ground truth depth map.

\subsection{Experimental Setup} \label{sec:setting}
In our experiments, we consistently adopt a fixed parameter setting for instance adjustment. In practice, we typically set the matching frequency threshold for redundancy pruning to 0 ane the $IoM = 0.7$ for duplication pruning.

For the learning rate of instance tokens during training, we typically set the coordinate learning rate $l_{xyz} = 2e{-3}$ and the scale learning rate $l_{s} = 2e{-3}$. When the scene contains a large number of instances with significant variations in scale, we appropriately increase these two parameters to better adapt to the more complex instance distributions. 

For a fair comparison, we consistently employ 2D semantic, instance, and panoptic segmentation images obtained through Grounded SAM as supervisory data for Kimera, PN, CL, PL, PR and PR++. Furthermore, to assess the impact of 2D VLM errors, we present an ablation study in the Appendix where we introduce different text prompts corresponding to objects that are unseen in the reconstructed scene.

PN requires high-quality 3D object polygonal bounding box ground truth aligned with 2D instance labels. On the three indoor datasets, such high-quality boxes are unavailable, so we provide them with object 3D rectangular bounding boxes obtained from instance mesh ground truth. For PVLFF, its default open-vocabulary segmentation methods are employed, which are compatible with its hierarchical clustering. 

We evaluate all frames of a scene but train on keyframes, ensuring each scene contains approximately 200-400 frames, to reduce computational costs and assess the results of novel views. The downsampling rate is adapted to the camera moving speed across scenes, but keeps all comparative methods the same. All experiments were conducted on a single NVIDIA A6000 GPU. Concurrently, we also provide more detailed numerical results and analysis of the computational efficiency in the Appendix.

\subsection{Scene-level Instance Segmentation} \label{exp:instance} 
We compare the scene-level instance segmentation performance of PanopticRecon++ with state-of-the-art panoptic and instance reconstruction systems in both 2D and 3D spaces. Different from the image-level instance segmentation, the scene-level instance segmentation requires the unique ID across images. 

\medskip \noindent\textbf{2D Instance Segmentation.}
Tab. \ref{tab:seg_replica}, \ref{tab:seg_scannet}, and \ref{tab:seg_scannet++} present quantitative results of both training and novel view instance segmentation comparison on the Replica, ScanNet-V2, and ScanNet++ datasets, respectively. Fig.~\ref{fig:seg_scannet} and Fig.~\ref{fig:seg_scannet++} show the visualization cases of the rendered segmentation images on the ScanNet-V2 and ScanNet++ datasets for comparison. 

As we can see, PanopticRecon++ demonstrates the best performance across all sequences and datasets, which owes to the cross-attention allowing for both spatial prior and direct back-propagation from instance loss. Compared with PN, which accesses the 3D ground truth, our methods also demonstrate better performance in most scenes, because PN is vulnerable to intersections between 3D rectangular bounding boxes in contrast to its original polygonal bounding boxes provided by KITTI-360. When dealing with scenes where objects have less intersection, "0628\_02" scene, our method is slightly weaker. Since PanopticRecon++ has no access to the 3D ground truth, this comparison validates the superior performance.

On Replica and ScanNet-v2, PR ranks after PanopticRecon++, because of the better prior brought by 3D space clustering than image space employed in PVLFF. However, PR performs much worse on ScanNet++, where PVLFF ranks second. The reason is also the clustering in PR that depends on the normals. The scene is under-segmented due to the similar normal directions, making the small instances in ScaNet++ connected to background stuff. As the method is two-stage, this error is accumulated, causing the degrading instance segmentation performance. Considering PanopticRecon++, of which the performance is leveraged by the 3D space prior and the end-to-end learning, avoiding the error accumulation in stages.

\begin{figure*}[t]
    \centering
    \includegraphics[width=\linewidth,keepaspectratio]{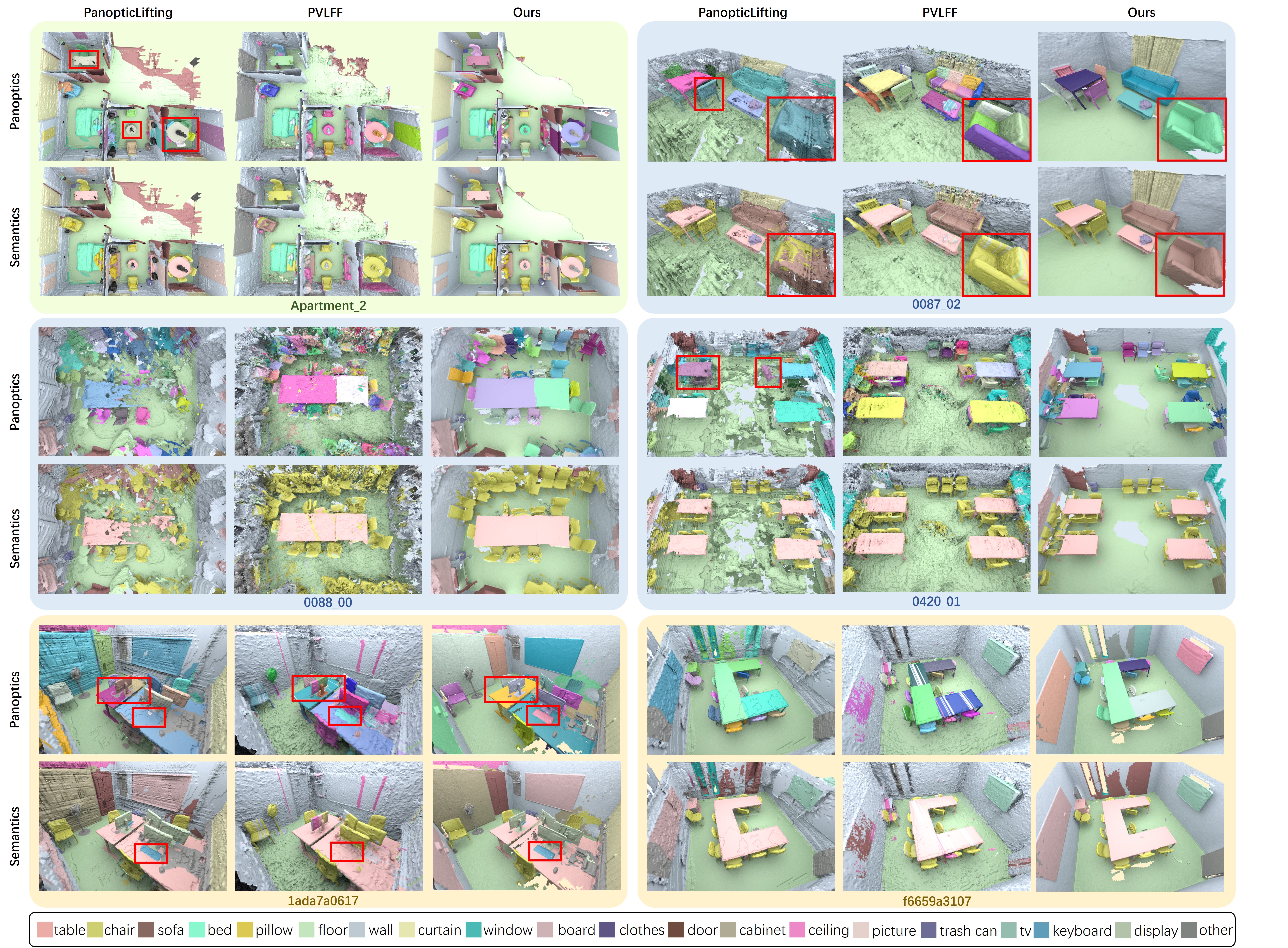}\\
    \vspace{-0.3cm}
    \caption{\label{Fig_mesh} Comparison of the panoptic and semantic reconstruction quality on ScanNet and ScanNet++ by different methods.}
\end{figure*}

Several cases are demonstrated in Fig.~\ref{Fig_mesh}, which show how the instance spatial priors help PanopticRecon++. 
A failure case from PL in "Apartment\_2" scene shows two chairs in different rooms are incorrectly grouped together, highlighting the limitations of image-level supervision in capturing spatial relationships. Incorporating spatial priors is necessary to rectify such errors.
In the "0420\_01" scene, the panoptic mesh from PL exhibits the same issue of multiple instances with identical colors (duplicated instance IDs), eg. one purple table and one purple chair near, even when these instances are spatially separated without any physical occlusion. This occurs due to limited observations covering both instances simultaneously.

\medskip \noindent\textbf{3D Instance Segmentation.} 
Compared with 2D instance segmentation, 3D instance segmentation is not affected by the distribution of image viewpoints. In addition, we query the ground truth point cloud for segmentation evaluation, in order to suppress the effect of reconstruction quality in the performance. 

PanopticRecon++ outperforms the other methods across all sequences of all datasets, demonstrating its effectiveness. However, in 3D, PanopticRecon++ is weaker than PN in most scenarios on mW-Cov (3D). The main reason is that the 3D instance segmentation is much more tightly coupled with the 3D bounding box than the 2D instance segmentation, and large instances have fewer intersections among their bounding boxes. Given the access to ground truth 3D bounding box, it is not surprising that PN has the best performance.

PR and PVLFF demonstrate a similar ranking on 3D instance segmentation as 2D instance segmentation.  
The only exception is "5748ce6f01" scene, where the performance of PVLFF drops. As shown in Fig.~\ref{fig:seg_scannet++}, this drop is attributed to the presence of a large, ambiguous instance (a TV screen with its stand) in the ground truth, leading to ambiguity in segmentation. Since there are few observations of this ambiguous instance in the sequence, this issue is not reflected in the 2D instance segmentation metrics but clarified in 3D performance. For PL, as shown in Tab.~\ref{tab:seg_scannet++} and Tab.~\ref{tab:seg_kitti}, the lack of spatial prior leads to multiple duplicate instance IDs assigned to objects that do not co-appear in the same frame, resulting in over-segmentation and inconsistent instance associations across views. This phenomenon directly contributes to its inferior instance segmentation performance. 

\subsection{Scene-level Semantic Segmentation} 
We evaluate the proposed method for comprehensive semantic scene segmentation in both 2D and 3D spaces. PanopticRecon++ and PN both integrate the instances cues for learning semantics. PL, CL, PVLFF, and PR learn a separate semantic field. Kimera is regarded as a reference to show the impact of the fusion strategy in semantic reconstruction.

\medskip \noindent\textbf{2D Semantic Segmentation.} 
As shown by the $mIoU$ and $mAcc$ in the 2D-Scene column of Tables \ref{tab:seg_replica}, \ref{tab:seg_scannet}, and \ref{tab:seg_scannet++}, our method perform the best on almost all sequences. We explain the advantage of the panoptic loss, which brings consistency between semantics and instances. As the semantics and instances are consistent in the observations, the panoptic loss brings additional constraints to guide the learning. In the "0087\_02" scene and "1ada7a0617" scene, PanopticRecon++ ranks second due to the distribution of the image viewpoint, where a mistake is frequently observed in 2D. PN also considers both, therefore, its semantic segmentation performance is correlated to the instance performance, resulting in the best performance in "0628\_02" scene. 

PL and CL employ the same semantic head. A soft constraint to enforce the unique class of an instance is applied, while in PanopticRecon++, the unique class of an instance is guaranteed by the architecture. For PR, no such instance level constraint is considered, thus showing inferior performance. Finally, for PVLFF, its inferior performance is mainly caused by two reasons: semantic prediction is supervised by a VLM independent from the instance VLM, and the prediction architecture is simply an inner product. Based on the analysis, the value of panoptic segmentation guided joint modeling and end-to-end learning in PanopticRecon++ is further verified.

In the "0087\_02" scene (Fig.~\ref{fig:seg_scannet}), both PL and CL mistakenly label the sofa on the right as a hybrid "chair" and "sofa" due to faulty 2D object detection from Grounded SAM. While PL accurately segments the sofa, the soft constraint fails to rectify the incorrect semantic label.

PL achieves the highest 2D semantic segmentation scores on KITTI-360 (Tab.~\ref{tab:seg_scannet++}), but only due to superior performance on two broad, ambiguously defined classes, \textit{building} and \textit{vegetation}, where its MLP-based smooth boundaries align with the coarse, smoothed ground truth in KITTI-360; for all other classes, our method significantly outperforms it.

\medskip \noindent\textbf{3D Semantic Segmentation.}
As shown in Tab.~\ref{tab:seg_scannet} and Tab.~\ref{tab:seg_scannet++}, our proposed method demonstrates a more pronounced advantage in instance-level semantic segmentation within the 3D space. As evidenced by the 3D semantic segmentation metrics, our method consistently outperforms state-of-the-art methods in terms of $mIoU$ (3D). We explain this result by the joint consideration of semantics and instance combined with 3D prior. While for $mAcc$ (3D), the slight decrease in $mAcc$ (3D) in 
"0420\_01" scene for inaccurate 3D semantic ground truth. 

PVLFF struggles with inconsistent VLM observations and feature similarity, leading to unsatisfactory performance in 2D semantic segmentation. PL, CL, and PR show varying performance rankings, attributed to the combined effects of viewpoint and class distributions. These factors minimally impact PanopticRecon++ in 2D versus 3D segmentation. We consider that the advantage is brought by the design of PanopticRecon++ on consistent prediction between semantics and instances, as well as points belonging to one instance.

\subsection{Scene-level Panoptic Segmentation} 
In this section, we evaluate the scene panoptic segmentation performance using $PQ$, $SQ$, and $RQ$. $PQ$, a metric that simultaneously reflects both segmentation and recognition quality, is the most indicative of panoptic segmentation performance. PR++ achieves the highest $PQ$ and $SQ$ scores across all sequences of indoor datasets, attributed to the end-to-end learning of panoptic segmentation. As shown in Fig.~\ref{fig:seg_scannet} and Fig.~\ref{fig:seg_scannet++}, only our method guarantees the consistency between semantic and instance labels, as they are derived by Bayes rule based parameter-free panoptic head. The softmax operation within the panoptic head enables simultaneous consideration of semantic and instance branches during inference and inherently prevents conflicts between the semantic label and instance ID. Only in "5748ce6f01" scene, PanopticRecon++ underperforms PN. PN benefits from the limited overlap ground truth bounding boxes in this sequence.
However, PN is provided with the highest-quality 3D polygonal bounding boxes on KITTI-360. As shown in Tab.~\ref{tab:seg_kitti}, this privileged access to precise priors enables PN to achieve the highest panoptic segmentation performance. In contrast, PanopticRecon++, operating without any 3D supervision, attains a PQ score competitive with PN.

In contrast, PL, CL, PVLFF, and PR methods query the instance branch based on foreground regions provided by the semantic branch, depending solely on semantic accuracy without considering the conflict between boundaries in semantic and instance segmentation heads, making the panoptic segmentation performance deteriorate. Note that in "5748ce6f01" scene, $RQ$ of PL beats PanopticRecon++, which is caused by the ambiguity of instance class definition in ground truth mentioned above. In summary, the significant improvement in panoptic segmentation performance validates the effectiveness of PanopticRecon++ in dealing with three challenges by differentiable assignment, spatial prior, and panoptic head. Several cases are shown in Fig.~\ref{fig:seg_scannet++}. In the "1ada7a0617" scene, only PanopticRecon++ achieves fully consistent semantic and instance masks for the keyboard on the desk.


\begin{table*}[!htbp]
    \centering
    \setlength{\tabcolsep}{4pt}
    \caption{Ablation Studies on ScanNet++ dataset}
    \vspace{-0.3cm}
    \centering
    \resizebox{\linewidth}{!}{
    \begin{tabular}{ccccc|ccc|cc|cc|cc|cc}
        \toprule
        \multicolumn{5}{c|}{Method} & \multicolumn{7}{c|}{2D-Scene} & \multicolumn{4}{c}{3D}\\
        \cmidrule(r){1-5}  \cmidrule(r){6-12} \cmidrule(r){13-16}
            3DG & Prior & Adj & Pan & $h_{SI}$
             & $\text{PQ}$$\uparrow$ & SQ$\uparrow$ & RQ$\uparrow$ 
             & mIoU$\uparrow$ & mAcc$\uparrow$ 
             & mCov$\uparrow$ & mW-Cov$\uparrow$ 
             & mIoU$\uparrow$ & mAcc$\uparrow$ 
             & mCov$\uparrow$ & mW-Cov$\uparrow$\\
        \midrule
            \ding{51} & \ding{51} & \ding{55} & \ding{55} & \ding{55}
            & 46.37 & 73.12 & 59.81 
            & 78.20 & 88.05 
            & 54.30 & 62.35  
            & 73.72 & 86.83
            & 67.48 & 65.17 \\
            \ding{51} & \ding{55} & \ding{51} & \ding{55} & \ding{55}
            & 65.73 & 73.26 & 82.40 
            & 71.61 & 81.10 
            & 56.29 & 68.71 
            & 64.98 & 75.94
            & 53.36 & 53.71 \\
            \ding{55} & \ding{51} & \ding{51} & \ding{55} & \ding{55}
            & 66.80 & 78.39 & 85.22 
            & 72.84 & 81.34 
            & 64.51 & 71.88 
            & 67.73 & 85.99
            & 72.46 & 70.69 \\
            
            \ding{51} & \ding{51} & \ding{51} & \ding{55} & \ding{55}
            & 74.63 & 82.42 & 89.78 
            & 79.78 & 90.21 
            & 71.67 & 76.68  
            & 74.21 & 87.49 
            & 84.25 & 76.51 \\ 

            \ding{51} & \ding{51} & \ding{51} & \ding{55} & \ding{51}
            & 74.63 & 82.42 & 89.78 
            & 78.21 & 89.44 
            & 71.67 & 76.68  
            & 70.81 & 84.58 
            & 84.25 & 76.51\\ 
            
            \ding{51} & \ding{51} & \ding{51} & \ding{51} & \ding{55}
            & \textbf{78.99} & \textbf{84.80} & \textbf{92.37} 
            & 80.64 & 90.37 
            & \textbf{76.01} & \textbf{79.12}  
            & 75.56 & 87.31
            & \textbf{88.98} & \textbf{79.16} \\
            \ding{51} & \ding{51} & \ding{51} & \ding{51} & \ding{51}
            & \textbf{78.99} & \textbf{84.80} & \textbf{92.37} 
            & \textbf{82.80} & \textbf{90.81} 
            & \textbf{76.01} & \textbf{79.12}  
            & \textbf{76.50} & \textbf{87.37}
            & \textbf{88.98} & \textbf{79.16} \\
        \bottomrule
    \end{tabular}
    }
\label{tab:ablation}
\vspace{-0.3cm}
\end{table*}

\begin{table}[t]
    \centering
    \setlength{\tabcolsep}{4pt}
    \caption{Reconstruction quality using different methods on 3 datasets}
    \vspace{-0.3cm}
    \resizebox{\linewidth}{!}{
    \begin{tabular}{c|l|ccc|ccc}
        \toprule
        Dataset & Method & Acc.$\downarrow$ & Com.$\downarrow$ & C-L1$\downarrow$ 
                                 & Pre.$\uparrow$ & Re.$\uparrow$ & F1$\uparrow$  \\
        \midrule
        \multirow{4}{*}{Replica} 
            & Kimera  
            & 1.98 & 8.39 & 5.19 
            & 88.13 & 75.06 & 81.07 \\
            & Panoptic Lifting 
            & 2.54 & 3.07 & 2.80 
            & 89.07 & 86.70 & 87.87 \\
            & PVLFF 
            & 3.78 & 5.75 & 4.76 
            & 77.98 & 68.05 & 72.68 \\
            & \textbf{PanopticRecon++} 
            & \textbf{1.06} & \textbf{2.29} & \textbf{1.68} 
            & \textbf{95.83} & \textbf{89.55} & \textbf{92.58} \\
        \midrule
        \multirow{4}{*}{ScanNet} 
            & Kimera  
            & 2.80 & 2.99 & 2.89 
            & 84.51 & 89.12 & 86.69 \\
            & Panoptic Lifting 
            & 6.00 & 6.90 & 6.45 
            & 51.41 & 55.42 & 53.29 \\
            & PVLFF 
            & 5.15 & 4.66 & 4.91 
            & 61.00 & 69.70 & 65.03 \\
            & \textbf{PanopticRecon++} 
            & \textbf{1.98} & \textbf{1.38} & \textbf{1.68} 
            & \textbf{91.77} & \textbf{94.82} & \textbf{93.20} \\
        \midrule
        \multirow{4}{*}{ScanNet++} 
            & Kimera  
            & 2.26 & 2.89 & 2.57 
            & 86.66 & 90.77 & 88.65 \\
            & Panoptic Lifting 
            & 3.98 & 3.80 & 3.89 
            & 72.76 & 79.44 & 75.90 \\
            & PVLFF 
            & 3.25 & 3.75 & 3.50 
            & 85.81 & 78.93 & 82.22 \\
            & \textbf{PanopticRecon++} 
            & \textbf{0.89} & \textbf{0.62} & \textbf{0.76} 
            & \textbf{99.97} & \textbf{99.02} & \textbf{99.59} \\
        \bottomrule
    \end{tabular}
    }
    \label{tab:recon_replica}
\end{table}
\subsection{Parameter Sensitivity}
\begin{figure}[t]
    \centering
    \includegraphics[width=\linewidth,keepaspectratio]{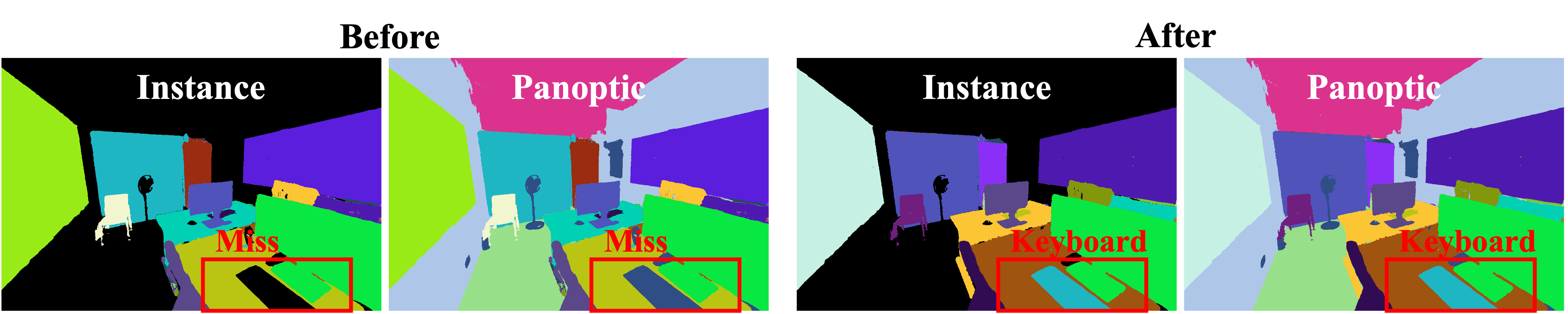}\\
    \caption{\label{Fig_param} Comparison of instance token learning rates before and after adjustment.}
    \vspace{-0.3cm}
\end{figure}

Our parameter settings are fairly robust for instances across most scenes; however, in certain special cases they exhibit some sensitivity and require appropriate adjustment. In practice, we typically adjust the learning rates of the instance tokens, including the coordinate learning rate $l_{xyz}$ and the scaling learning rate $l_{s}$. These parameters determine the movement speed of instance tokens during training as well as their spatial influence. For example, in scene 1ada7a0617 from ScanNet++, using the commonly adopted learning-rate settings (mentioned in Sec. \ref{sec:setting}) results in the keyboard on the table not being matched by any instance token, as shown in Fig. \ref{Fig_param}. To address this, we set $l_{xyz}= 5e{-3}$ and $l_{s}= 1e{-2}$ to facilitate faster matching of instance tokens to instances of different scales. 


\subsection{Geometric Reconstruction} \label{section:recon}
To evaluate the quality of scene geometric reconstruction, we select Kimera, PL, and PVLFF as our baseline methods and conduct experiments on three datasets: Replica, ScanNet-V2, and ScanNet++. CL and PR were not chosen as baselines due to their geometric reconstruction approaches being the same as PL and PR++, respectively. Notably, both Replica and ScanNet++ provide depth observations projected from high-precision scene mesh models, while ScanNet-V2 offers noisy depth maps obtained from real depth cameras, and its mesh ground truth is of relatively lower quality. Consequently, the reconstruction quality of all methods on ScanNet-V2 is generally lower compared to the other two datasets.
As shown in Tab.~\ref{tab:recon_replica}, our proposed method achieves reconstruction accuracy less than $1cm$ on both Replica and ScanNet++, and less than $3cm$ on ScanNet-V2, significantly outperforming other methods. In contrast to traditional reconstruction methods like Voxblox \cite{oleynikova2017voxblox} employed in Kimera, our approach leverages a multi-level feature grid and a compact MLP to represent the SDF surface, enhancing surface smoothness and completeness through regularization terms. While our method places a greater emphasis on surface details, it does not prioritize real-time performance, potentially leading to longer optimization times compared to Voxblox.
Both PL and PVLFF are implicit reconstruction methods that rely on density-based representations, which are prone to generating floating spatial artifacts and are less focused on surface quality compared to SDF-based methods. Therefore, considering both panoptic segmentation and reconstruction, the PanopticRecon++ contributes even better performance than the 3D segmentation metrics in Tab.~\ref{tab:seg_replica}, Tab.~\ref{tab:seg_scannet} and Tab.~\ref{tab:seg_scannet++}. Several meshes colorized by panoptic classes and semantic classes are visualized in Fig.~\ref{Fig_mesh}. As shown in cases, PanopticRecon++ produces smoother wall and floor surfaces, as well as more detailed small objects (e.g. keyboard and mouse).

\subsection{Ablation Studies} \label{section:ablation}
We perform ablation studies of segmentation and rendering by the average performance of 3 sequences in ScanNet++.



We analyze the five key design choices of PanopticRecon++: 3D Gaussian-modulated token representation (3DG), instance spatial prior (Prior), dynamic adjustment for tokens (Adj), panoptic head (Pan), and instance-level class feature ($h_{SI}$). The results of ablation studies are shown in Tab. \ref{tab:ablation}. 

\medskip \noindent\textbf{Token Representation.}
Table~\ref{tab:ablation} shows that 3D Gaussian-modulated tokens (3DG) outperform point cloud-based instance tokens for segmentation (rows 4 vs. 3). This is because point cloud-based tokens, which rely on Euclidean distances, poorly represent objects with significant scale and geometric variation, such as poles, keyboards, and elongated tables. Euclidean distance inadequately captures instances' spatial extent, orientation, and structure.
In contrast, the 3D Gaussian-modulated token explicitly learns the spatial distribution of each instance, naturally accommodating differences in scale and shape. This leads to a more principled and discriminative modeling of inter-instance geometric relationships, resulting in significantly improved segmentation performance. 

\medskip \noindent\textbf{Spatial Prior.}
The comparison between the second and fourth rows in Tab.~\ref{tab:ablation} clearly demonstrates the critical contribution of spatial prior (Eq.~\ref{eq:wight}) to improved segmentation performance. Interestingly, the result in the second row closely matches the segmentation performance of PL method. This indicates that relying solely on 2D supervision can result in different objects being mistakenly assigned the same instance ID, especially for objects that do not co-occur in the same image, leading to a substantial decline in segmentation performance.

\medskip \noindent\textbf{Dynamic Adjustment.}
As demonstrated by the comparison between the first and fourth rows in Tab.~\ref{tab:ablation}, we found that the lack of dynamic adjustment results in over-segmentation, significantly degrading the segmentation performance. This is primarily attributed to the redundancy and overlap of independently optimized binary masks, leading to too many instances.
When dynamic adjustment is applied solely, the removal of overlapping and intersecting binary masks results in a notable performance boost. 
By combining instance spatial prior with dynamic adjustment, our method reveals the advantages of instance tokens and spatial prior, leading to significant improvements in panoptic and instance segmentation performance.

\medskip \noindent\textbf{Panoptic Head and Instance-level Class.}
The spatial prior and dynamic token adjustment are indispensable; removing them leads to a severe performance drop (Tab. \ref{tab:ablation}, rows 1–2). The panoptic head and the instance-level class feature $h_{SI}$, in contrast, provide complementary enhancements. To evaluate their contributions, we fix the prior and adjustment and compare settings with/without the panoptic head and $h_{SI}$. Results indicate that $h_{SI}$ effectively refines noisy semantic predictions when instance segmentation is reliable, but misaligned tokens propagate errors, yielding performance even inferior to supervision from 2D semantics alone.
Additionally, both experiments demonstrate an improvement in $PQ$ compared to the model without the panoptic head, which can be attributed to the panoptic head's capability of Bayes rule based semantic and instance integration.

\subsection{User Case}
We demonstrate the feasibility of our system for robot simulation. We load the meshes generated from models trained on three scenes of ScanNet++ into Gazebo to ensure physical interaction between the robot and environment. Given goals in Gazebo, the robot, Jackal UGV, starts navigation. During the traversal, Gazebo engine integrates the camera poses, upon which the realistic RGB images, depth images, and panoptic segmentation images are rendered by novel view synthesis using PanopticRecon++. These views can be regarded as the robot sensor inputs to activate the algorithms in the simulation. Please refer to the multimedia material on project homepage.

\subsection{Failure Case and Limitation}
\begin{figure}[t]
    \centering
    \includegraphics[width=\linewidth,keepaspectratio]{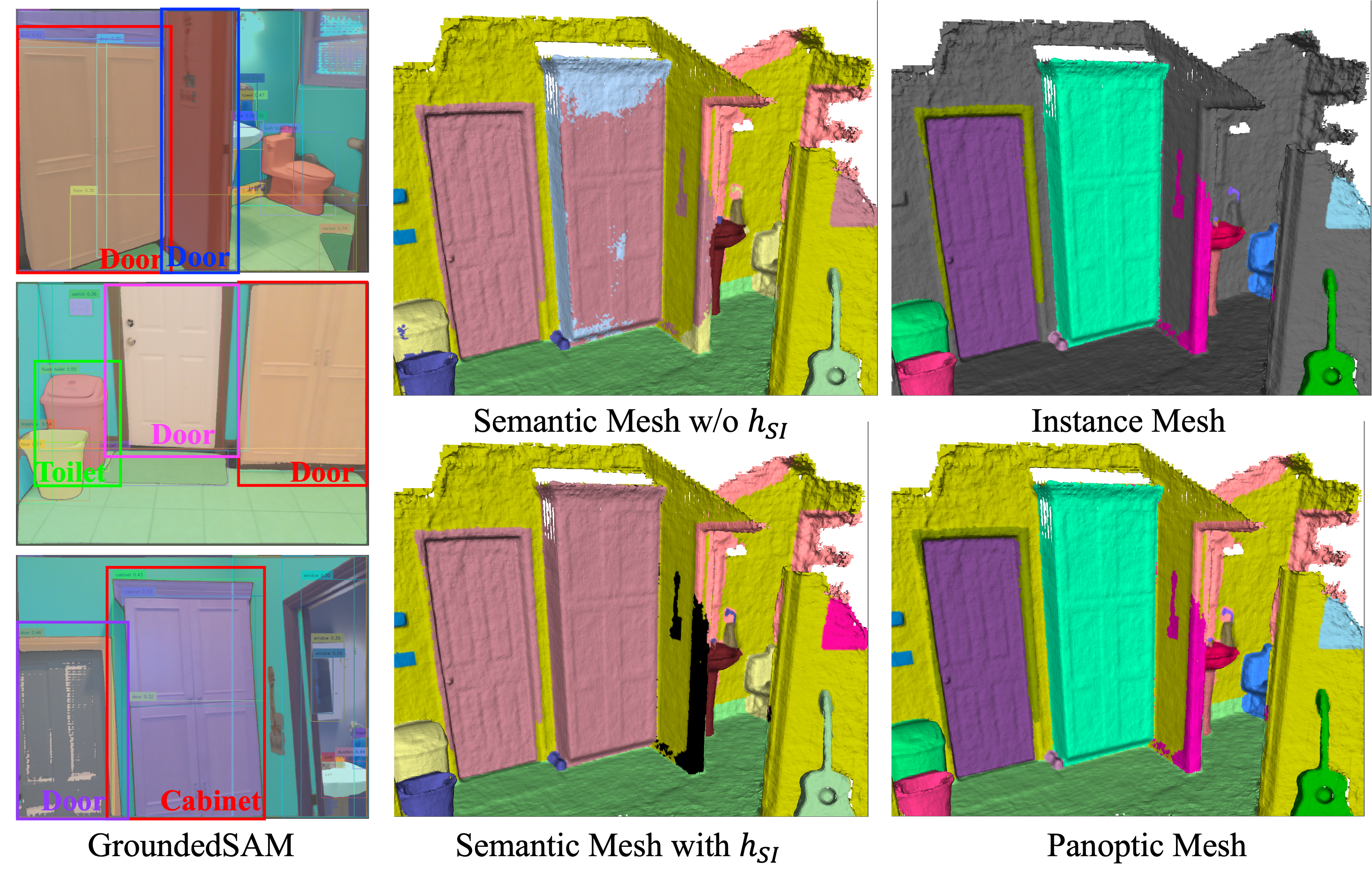}\\
    \vspace{-0.3cm}
    \caption{\label{Fig_failurecase}Failure Case: 3D Segmentation Errors Caused by GroundedSAM Mislabeling}
    \vspace{-0.3cm}
\end{figure}

As shown in Fig.~\ref{Fig_failurecase}, the limited generalization of 2D foundation models sometimes restricts our system. On the left, GroundedSAM exhibits typical errors: the same cabinet is labeled inconsistently (``cabinet'' / ``door''), a door is segmented with varying granularity due to its frame, and the trash bin is misclassified as a toilet. On the right, our method shows robustness to granularity variation (e.g., consistent cabinet instance), but frequent semantic mislabeling still propagates into instance tokens. As a result, although fragmented 2D labels are alleviated, the entire instance may still be assigned an incorrect semantic label. 

Another limitation of our method is the high GPU memory consumption due to the large scene containing many objects. This is because the memory cost of our attention mechanism scales with the number of instances.

\section{Conclusion}
We propose PanopticRecon++, an end-to-end open-vocabulary panoptic reconstruction method with multi-branch neural fields and 3D Gaussian-modulated instance tokens. Through cross-attention, segmentation features are integrated with spatial priors, enabling token coordinate back-propagation and resolving 3D instance ambiguity without post-processing. Learnable tokens allow controlled initialization and dynamic adjustment, while a parameter-free panoptic head ensures consistent semantic and instance labels. Experiments on simulated and real datasets show that PanopticRecon++ achieves globally consistent 3D segmentation and reconstruction with competitive performance. 
In future work, we plan to distill the entire scene into 3DGS to achieve real-time performance for robotic systems.

\bibliographystyle{IEEEtran}
\bibliography{IEEEabrv,bibliography}



 
\vspace{3pt}


\begin{IEEEbiography}[{\includegraphics[width=0.8in,height=1in,clip]{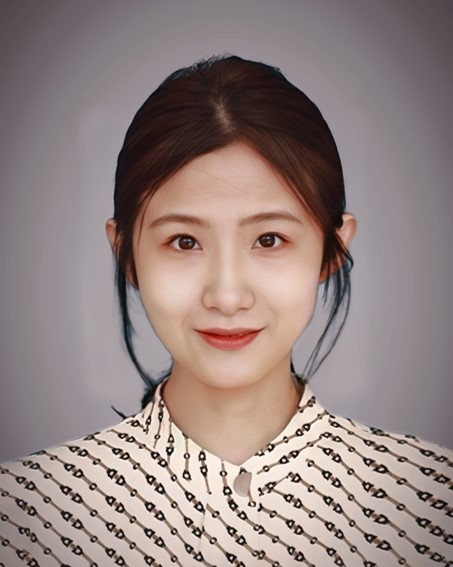}}]{Xuan Yu}
received her B.Eng. degree in Mechanical, Electrical and Information Engineering from Shandong University, China in 2021. She is currently a Ph.D. candidate in the Department of Control Science and Engineering at Zhejiang University. Her research interests include 3D vision and scene understanding.
\end{IEEEbiography}

\begin{IEEEbiography}[{\includegraphics[width=0.8in,height=1in,clip,keepaspectratio]{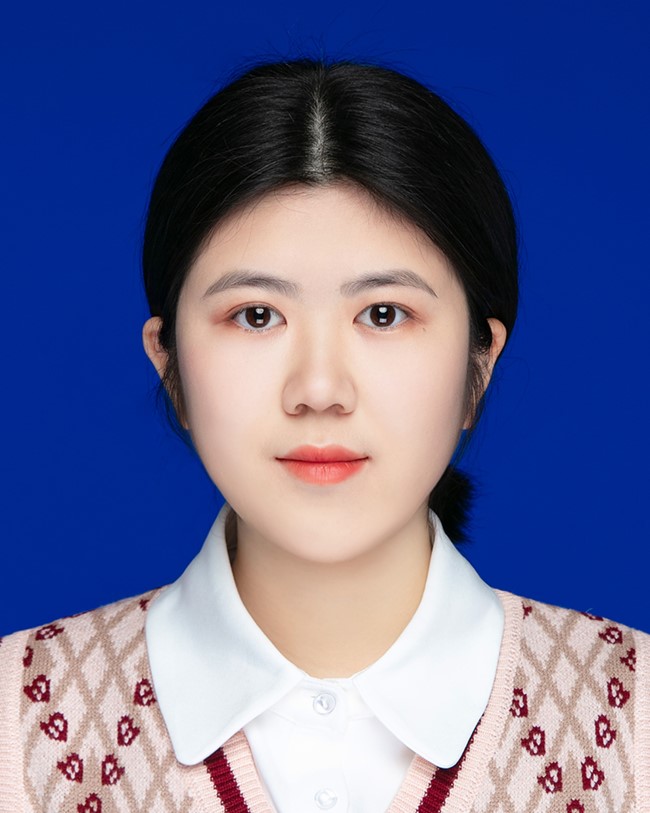}}]{Yuxuan Xie}
received her B.Eng. degree in Artificial Intelligence from Tongji University, China in 2024. She is currently pursuing her M.S. degree in the Department of Control Science and Engineering at Zhejiang University. Her research interests include 3D reconstruction and robotics.
\end{IEEEbiography}

\begin{IEEEbiography}[{\includegraphics[width=0.8in,height=1in,clip,keepaspectratio]{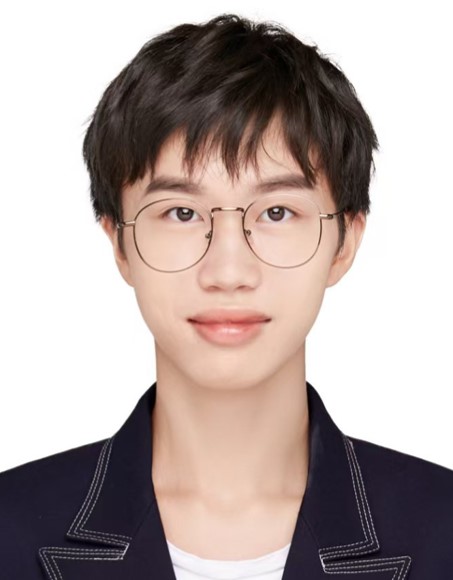}}]{Yili Liu}
received the B.Eng. degree in robotics engineering from Zhejiang University, China in 2023, where he is currently persuing the M.S. degree. His research interests include 3D perception and reconstruction on robotics and autonomous driving.
\end{IEEEbiography}

\begin{IEEEbiography}[{\includegraphics[width=0.8in,height=1in,clip,keepaspectratio]{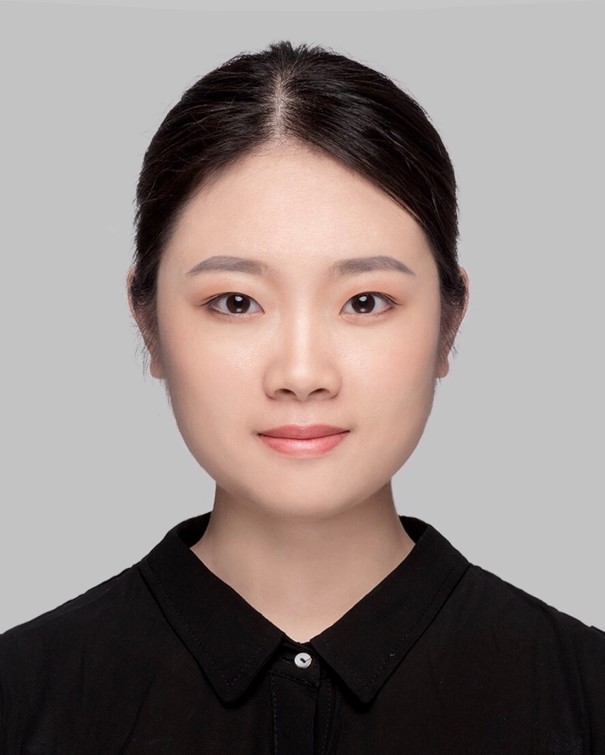}}]{Sitong Mao}
received the B.Eng. degree in computer science, Xiamen University, China in 2015, and the Ph.D. degree from the Department of Computing, Hong Kong Polytechnic University, China in 2021. She is currently a research scientist in the Embodied AI Innovation Lab, Huawei, Shenzhen, China. Her research interests include CV, Multi-modality, Embodied AI, and Transfer learning.
\end{IEEEbiography}

\begin{IEEEbiography}[{\includegraphics[width=0.8in,height=1in,clip,keepaspectratio]{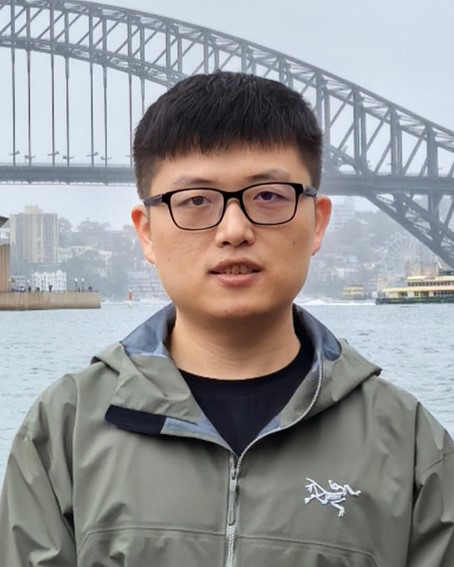}}]{Shunbo Zhou}
received his Ph.D. degree from the Department of Mechanical and Automation Engineering, Chinese University of Hong Kong, China in 2020. He is currently a research scientist in the Embodied AI Innovation Lab, Huawei, Shenzhen, China. His latest research interests include autonomous systems and embodied intelligence.
\end{IEEEbiography}

\begin{IEEEbiography}[{\includegraphics[width=0.8in,height=1in,clip]{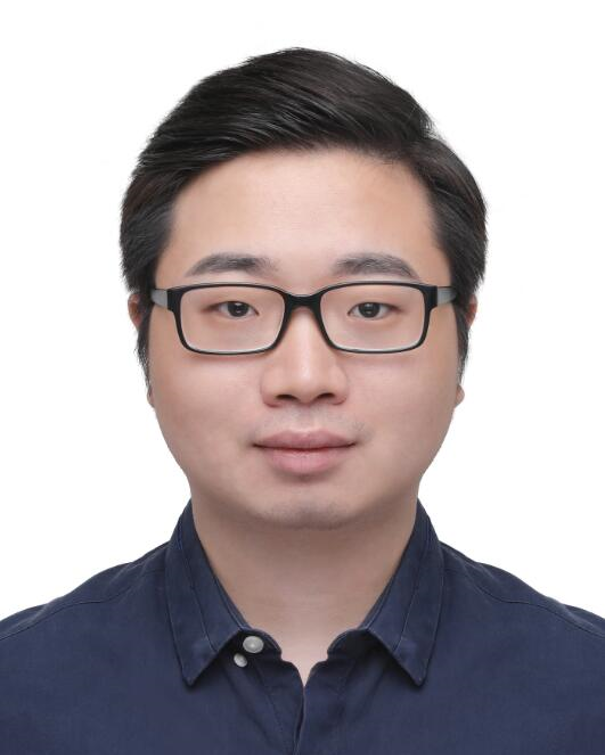}}]{Haojian Lu}
(Member, IEEE) received Ph.D. degree in Robotics from City University of Hong Kong in 2019. He was a Research Assistant at City University of Hong Kong, from 2019 to 2020. He is currently a professor in the Department of Control Science and Engineering, Zhejiang University. His research interests include bioinspired robotics, medical robotics, and soft robotics.
\end{IEEEbiography}

\begin{IEEEbiography}[{\includegraphics[width=0.8in,height=1in,clip]{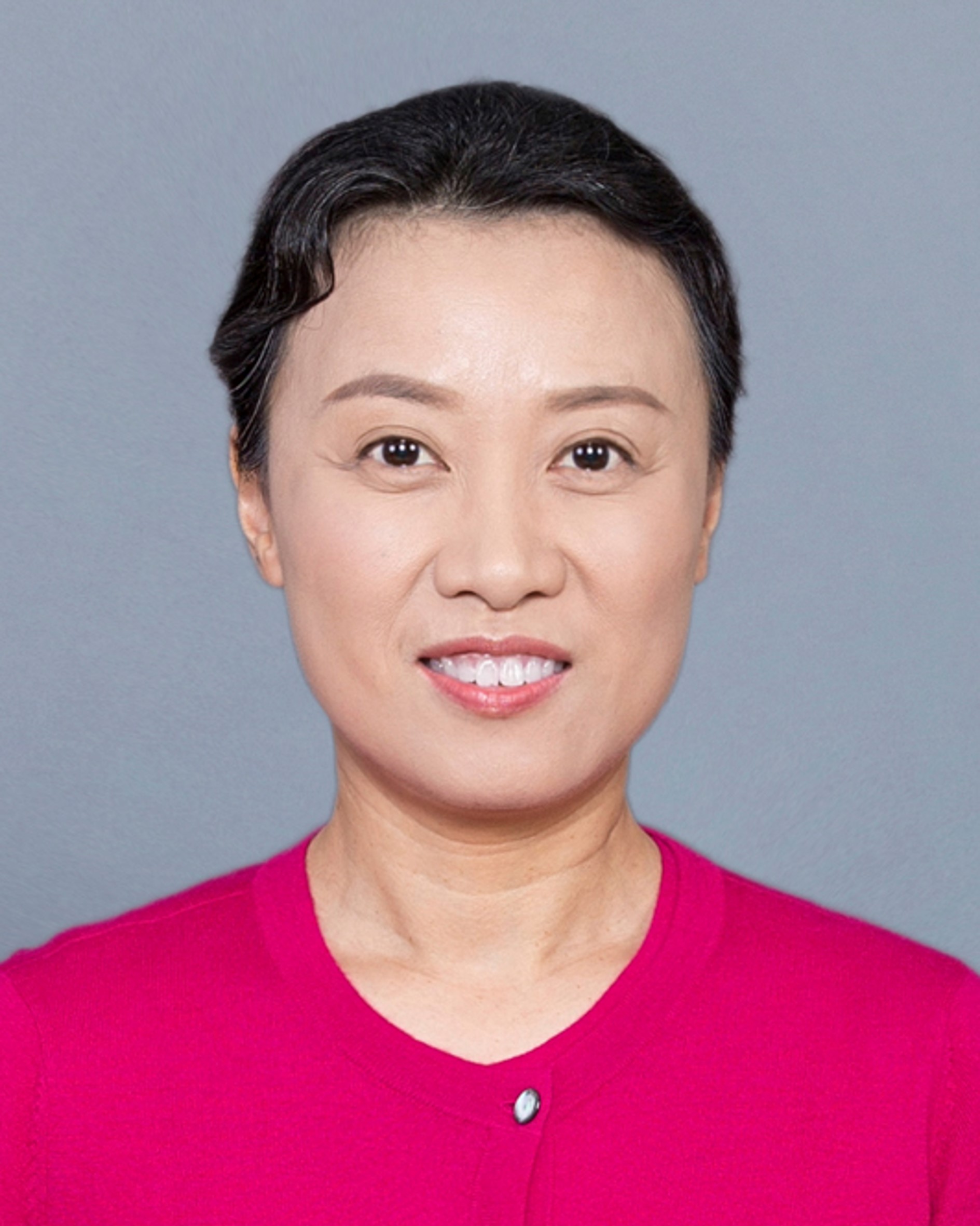}}]{Rong Xiong}
received her PhD in Control Science and Engineering from the Department of Control Science and Engineering, Zhejiang University, China in 2009. She is currently a Professor in the Department of Control Science and Engineering, Zhejiang University, China. Her latest research interests include embodied intelligence and motion control.
\end{IEEEbiography}

\begin{IEEEbiography}[{\includegraphics[width=0.8in,height=1in,clip]{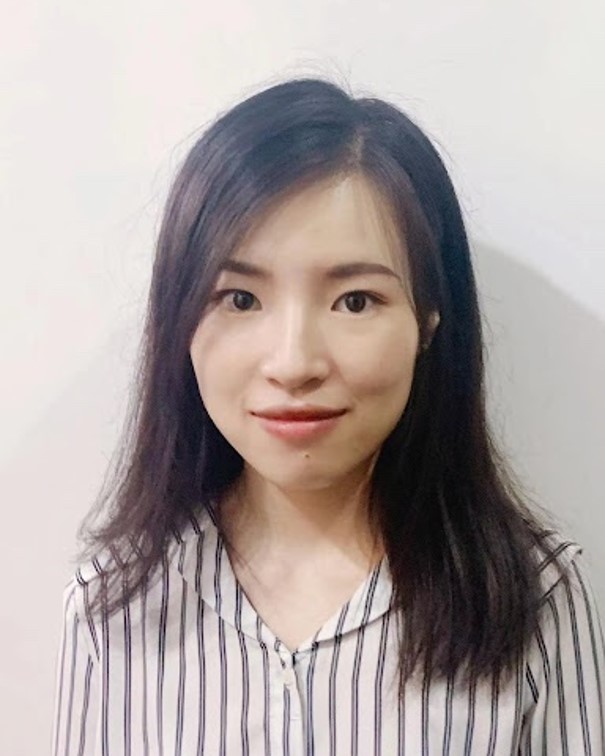}}]{Yiyi Liao}
received her Ph.D. degree from the Department of Control Science and Engineering, Zhejiang University, China in 2018. From 2018 to 2021, she was a postdoctoral researcher at the Autonomous Vision Group, University of Tübingen and Max Planck Institute for Intelligent
Systems, Germany. She is currently an Assistant Professor at Zhejiang University. Her research interests include 3D vision, scene understanding and 3D generative models.
\end{IEEEbiography}

\begin{IEEEbiography}[{\includegraphics[width=0.8in,height=1in,clip]{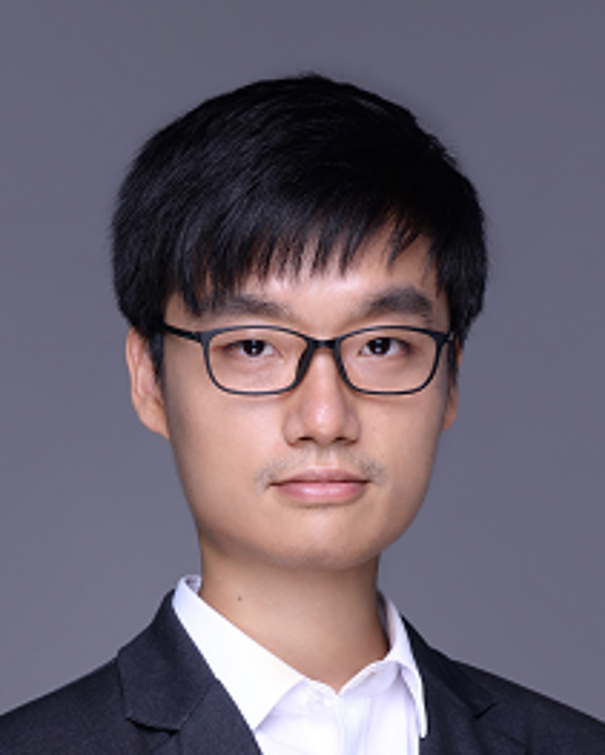}}]{Yue Wang}
received his Ph.D. degree in Control Science and Engineering from Department of Control Science and Engineering, Zhejiang University, China in 2016. He is currently a Professor in the Department of Control Science and Engineering, Zhejiang University, China. His
latest research interests include mobile robotics and robot perception.
\end{IEEEbiography}



\clearpage
\onecolumn
\begin{center}
    \textbf{\LARGE Appendix} 
\end{center}


\subsection{Generalization to Large-scale Scenes}

\begin{table}[H]
    \centering
    \setlength{\tabcolsep}{4pt}
    \caption{Panoptic Segmentation quality using different methods on KITTI-360 dataset}
    \centering
    \begin{threeparttable}
    \vspace{-0.3cm}
    \resizebox{\linewidth}{!}{
    \begin{tabular}{lccc|cc|cc|cc|cc}
        \toprule
            \multirow{2}*{Method} & \multicolumn{7}{c|}{2D-Scene} & \multicolumn{4}{c}{3D}\\
            \cmidrule(r){2-8} \cmidrule(r){9-12}
             & $\text{PQ}$$\uparrow$ & SQ$\uparrow$ & RQ$\uparrow$ 
             & mIoU$\uparrow$ & mAcc$\uparrow$ 
             & mCov$\uparrow$ & mW-Cov$\uparrow$ 
             & mIoU$\uparrow$ & mAcc$\uparrow$ 
             & mCov$\uparrow$ & mW-Cov$\uparrow$ \\

        \midrule
        \textit{Close-vocabulary:} \\
            Panoptic NeRF
            & \textbf{62.26} & \textbf{68.97} & \textbf{78.16} 
            & \textbf{64.39} & \textbf{74.59}  
            & \textbf{74.99} & \textbf{91.96}  
            & 61.24 & 69.91 
            & 88.05 & \textbf{92.30} \\ 
        \cmidrule(r){2-12}
        \textit{Open-vocabulary:} \\
            Panoptic Lifting
            & 42.50 & 54.79 & 55.33 
            & \cellcolor{red!30}\textbf{58.06} & \cellcolor{red!30}\textbf{74.20} 
            & 28.86 & 41.99 
            & 54.16 & 72.93 
            & 21.93 & 29.44 \\ 
            PVLFF 
            & 45.20 & 52.76 & 57.69 
            & 54.81 & 61.69 
            & 13.33 & 23.89 
            & 26.69 & 36.65 
            & 13.59 & 15.87 \\ 
            PanopticRecon 
            & \cellcolor{orange!30}57.61 & \cellcolor{orange!30}66.51 & \cellcolor{orange!30}74.16 
            & 56.21 & 69.86 
            & \cellcolor{orange!30}64.42 & \cellcolor{orange!30}85.98 
            & \cellcolor{orange!30}62.60 & \cellcolor{orange!30}80.59 
            & \cellcolor{orange!30}86.36 & \cellcolor{orange!30}85.99 \\ 
            \textbf{PanopticRecon++} 
            & \cellcolor{red!30}\textbf{60.23} & \cellcolor{red!30}\textbf{68.34} & \cellcolor{red!30}\textbf{74.51} 
            & \cellcolor{orange!30}57.36 & \cellcolor{orange!30}71.66 
            & \cellcolor{red!30}\textbf{68.59} & \cellcolor{red!30}\textbf{86.39} 
            & \cellcolor{red!30}\textbf{67.89} & \cellcolor{red!30}\textbf{82.42} 
            & \cellcolor{red!30}\textbf{91.26} & \cellcolor{red!30}\textbf{87.46} \\ 
            \bottomrule
        \end{tabular}
    }
    \end{threeparttable}
    \label{tab:seg_kitti}
\end{table}

\begin{figure}[H]
    \centering
    \includegraphics[width=\linewidth,keepaspectratio]{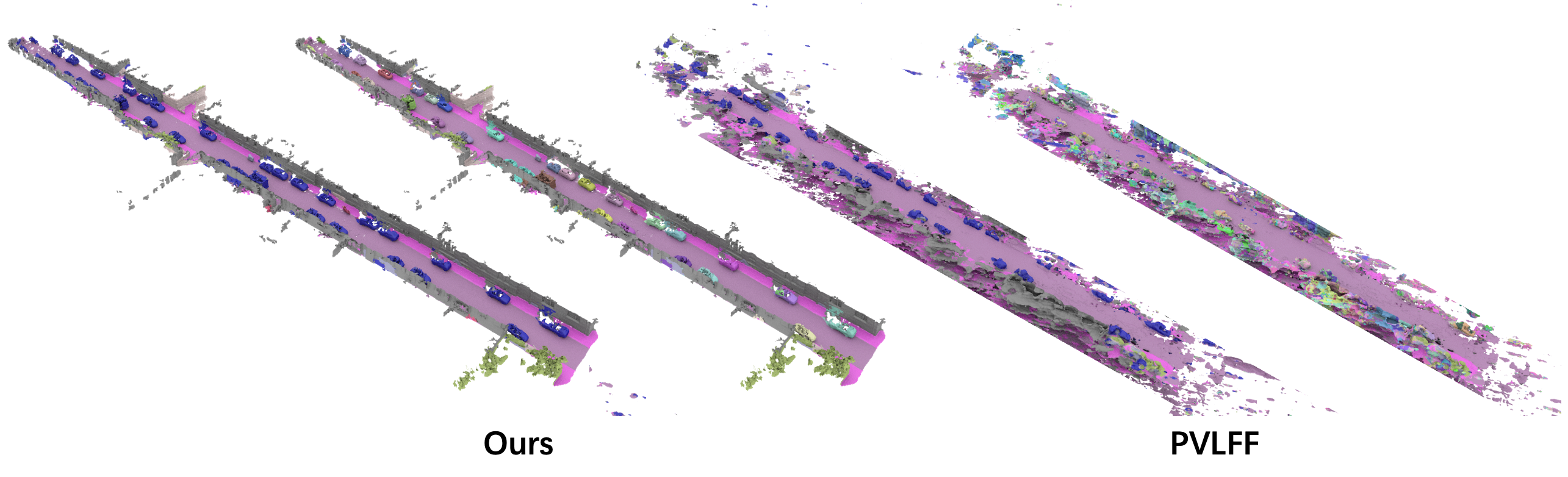}\\
    \caption{\label{fig:bicycle} Comparison of the quality of semantic segmentation and panoptic meshes of different methods on KITTI-360.}
\end{figure} %

\begin{figure}[H]
    \centering
    \includegraphics[width=0.8\linewidth,keepaspectratio]{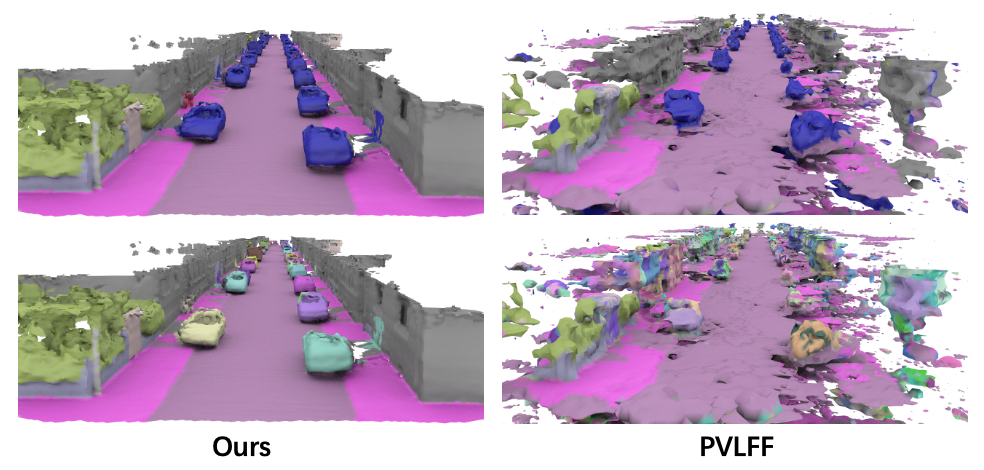}\\
    \caption{\label{fig:bicycle} Comparison of the quality of semantic segmentation and panoptic meshes of different methods on KITTI-360.}
\end{figure} %

To validate the generalization ability of our method across varying scene scales and sensor modalities, we extend our evaluation to outdoor large-scale scenes, utilizing RGB images and LiDAR point clouds as joint multimodal inputs. Specifically, we select a 200m segment of sequence 00 from the KITTI-360 dataset for this evaluation. The sensor configuration includes depth observations from a Velodyne HDL-64 LiDAR and a single perspective camera. In our experiments, the set of semantic labels comprises: \textit{road}, \textit{sidewalk}, \textit{building}, \textit{fence}, \textit{vegetation}, \textit{car}, \textit{bicycle}, and \textit{motorcycle}. Notably, \textit{car} and \textit{motorcycle} are designated as instance-level categories.

As shown in Tab. \ref{tab:seg_kitti}, Panoptic NeRF, as the closed-vocabulary approach, achieves the highest panoptic segmentation performance, as it is trained using known 3D bounding boxes and precisely aligned semantic classes, effectively operating under the assumption of perfect instance ground truth. Consequently, its results serve as a strong upper bound on achievable segmentation accuracy. Our open-vocabulary method achieves a PQ of 60.23\%, just 2\% shy of Panoptic NeRF's performance. 

Crucially, the primary advantage of open-vocabulary methods lies in their ability to segment objects absent from the training set and to produce finer, more accurate segmentation boundaries. For instance, as shown in Fig. \ref{fig:bicycle}, the bicycle was omitted from the ground-truth annotations; however, the 2D VLM successfully detected it, enabling our open-vocabulary framework to recover its shape and spatial extent accurately.

\begin{figure}[!t]
    \centering
    \includegraphics[width=\linewidth,keepaspectratio]{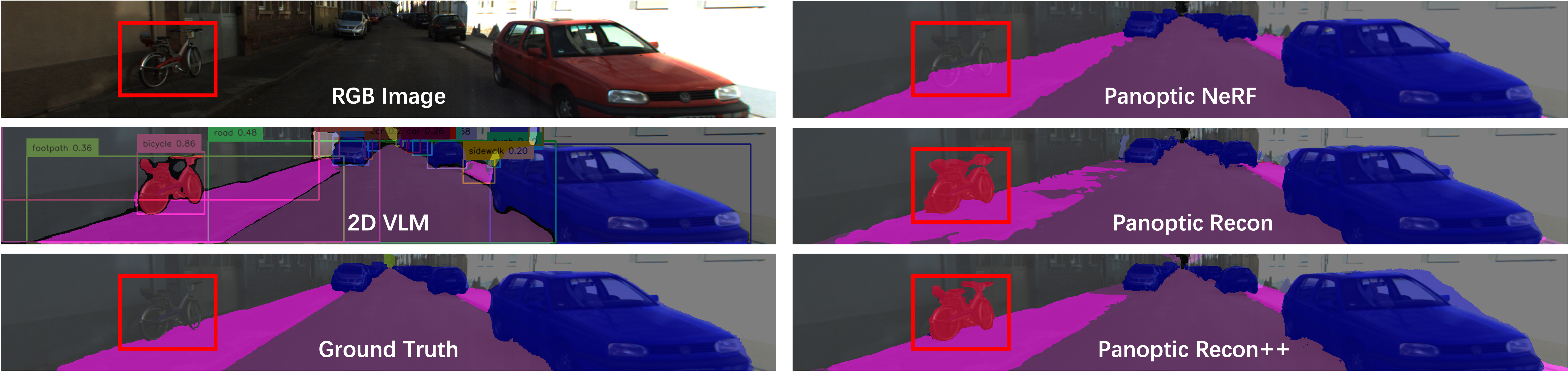}\\
    \caption{\label{fig:bicycle} Comparison of the quality of semantic segmentation and panoptic meshes of different methods on KITTI-360.}
\end{figure} %

Among panoptic segmentation metrics, our method achieves the highest PQ, while Panoptic Lifting attains the highest semantic segmentation metrics. This discrepancy arises because, among all semantic classes, only \textit{building} and \textit{vegetation} achieve higher segmentation performance than our method. In particular, \textit{vegetation} shows a significant performance advantage over ours. For all other classes, Panoptic Lifting performs worse than our approach. This behavior stems from Panoptic Lifting’s reliance on an MLP-based representation, which produces smoother, more regularized boundaries. Furthermore, the ground-truth annotations in KITTI-360 are themselves smoothed and less precise than the true object boundaries in the raw observations. Consequently, Panoptic Lifting benefits from this alignment between its inherent smoothness and the annotated ground truth, particularly for large, ambiguously bounded regions such as \textit{vegetation} and \textit{building}, leading to superior performance on these two categories.

\subsection{Impact of 2D VLM Errors}
\textbf{Evaluate on unseen categories}:  
We introduce an additional ablation study to evaluate the influence of different text prompt lists on the segmentation performance of our method. Specifically, we conduct experiments on the ScanNet scene 0087\_02 using three distinct text prompt configurations: (a) a baseline prompt list augmented with \textit{plant}; (b) an extended list including \textit{plant} and \textit{box}; (c) an extended list further including \textit{plant}, \textit{box}, and \textit{shelf}; and (d) the original prompt list from our manuscript, serving as the baseline. This comparison enables a systematic analysis of how vocabulary expansion affects open-vocabulary segmentation accuracy and robustness.

\begin{table}[H]
    \centering
    \setlength{\tabcolsep}{4pt}
    \caption{Evaluate on Unseen Categories with Different Text Prompt Lists}
    \vspace{-0.3cm}
    \centering
    \resizebox{\linewidth}{!}{
    \begin{tabular}{c|ccc|cc|cc|cc|cc}
        \toprule
        \multirow{2}*{Method} & \multicolumn{7}{c|}{2D-Scene} & \multicolumn{4}{c}{3D}\\
            \cmidrule(r){2-8} \cmidrule(r){9-12}
             & $\text{PQ}$$\uparrow$ & SQ$\uparrow$ & RQ$\uparrow$ 
             & mIoU$\uparrow$ & mAcc$\uparrow$ 
             & mCov$\uparrow$ & mW-Cov$\uparrow$ 
             & mIoU$\uparrow$ & mAcc$\uparrow$ 
             & mCov$\uparrow$ & mW-Cov$\uparrow$\\
        \midrule
            a
            & \textbf{81.84} & \textbf{88.71} & \textbf{92.38} 
            & \textbf{81.49} & \textbf{87.96} 
            & \textbf{84.81} & \textbf{90.54}  
            & \textbf{80.27} & \textbf{87.28}
            & \textbf{98.24} & \textbf{86.65} \\
            b 
            & 79.64 & 88.34 & 90.15 
            & 81.02 & 86.59 
            & 84.21 & 89.48  
            & 78.50 & 82.88
            & 94.63 & 86.34 \\
            c 
            & 77.52 & 77.52 & 87.50 
            & 74.39 & 79.51 
            & 83.20 & 89.33  
            & 71.41 & 80.39
            & 89.28 & 85.30 \\
            d
            & \textbf{81.84} & \textbf{88.71} & \textbf{92.38} 
            & \textbf{81.49} & \textbf{87.96} 
            & \textbf{84.81} & \textbf{90.54}  
            & \textbf{80.27} & \textbf{87.28}
            & \textbf{98.24} & \textbf{86.65} \\
        \bottomrule
    \end{tabular}
    }
\label{tab:ablation2}
\end{table}

Since the scene 0087\_02 contains no instances of \textit{plant} nor any semantically similar objects, the addition of \textit{plant} to the text prompt list had no measurable impact on the 2D VLM’s mask predictions, resulting in identical segmentation performance between (a) and (d) as shown in Tab \ref{tab:ablation2}. 
In contrast, (b) introduces an additional prompt \textit{box}, which triggers only minor false positives in a small subset of images where partial views of tables are observed, resulting in a marginal degradation of segmentation metrics.
(c) further incorporates the \textit{shelf} prompt in addition to (b), increasing misclassifications: in numerous images where doors are partially occluded or incompletely observed, the 2D vision-language model (VLM) erroneously assigns \textit{shelf} labels due to structural similarities between door frames and shelf geometries. This proliferation of false positives compromises the accuracy of the segmentation output, resulting in a drop in overall metrics compared to (b).
These results demonstrate that while text prompts for 2D VLMs should be carefully curated to maximize segmentation fidelity, the system remains robust even under suboptimal prompt selection.

\begin{table}[H]
    \centering
    \setlength{\tabcolsep}{4pt}
    \caption{Evaluate on Unseen Categories with Different Text Prompt Lists}
    \vspace{-0.3cm}
    \centering
    \resizebox{0.5\linewidth}{!}{
    \begin{tabular}{c|ccc}
        \toprule
             & $\text{Model Size}$$\downarrow$ & Inference Time$\downarrow$ & Memory$\downarrow$ \\
        \midrule
            Panoptic Lifting 
            & 35M & 14.5s & 15G \\
            PVLFF
            & 110M & 12s & 10G \\
            Panoptic Recon++
            & 900M & 4s & 30G \\
        \bottomrule
    \end{tabular}
    }
\label{tab:efficiency}
\vspace{-0.3cm}
\end{table}

\subsection{Computational Efficiency}
We evaluate the model size, per-frame inference time, and memory consumption of our method and competing approaches directly on KITTI-360 using an NVIDIA A6000 GPU. All methods are tested on images of resolution 376×1408. 
As shown in Tab. \ref{tab:efficiency}, Panoptic Lifting is based on MLP representations, which offer compact model sizes but suffer from slow inference. Both PVLFF and Panoptic Recon++ employ implicit representations over grid-based structures; however, PVLFF incurs additional inference overhead due to costly feature clustering operations. In contrast, our method achieves a substantial acceleration in inference speed, at the cost of a moderate increase in model size.

\begin{figure}[H]
    \centering
    \includegraphics[width=\linewidth,keepaspectratio]{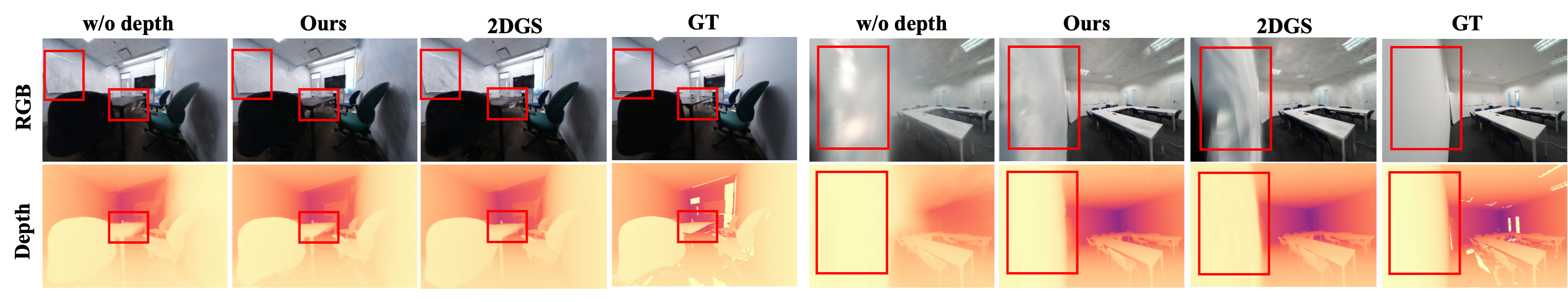}\\
    \vspace{-0.3cm}
    \caption{\label{Fig_3dgs} The ablation of $L_d$ in (\ref{eq:rgb_loss}) on novel views of ScanNet++ dataset. the first row shows the rendered RGB images and their corresponding rendered depth images are shown on the second row. Rendering details are highlighted in the red boxes.}
\end{figure} 

\subsection{Ablation Study on Rendering}
\begin{figure}[H]
    \centering
    \includegraphics[width=0.5\linewidth,keepaspectratio]{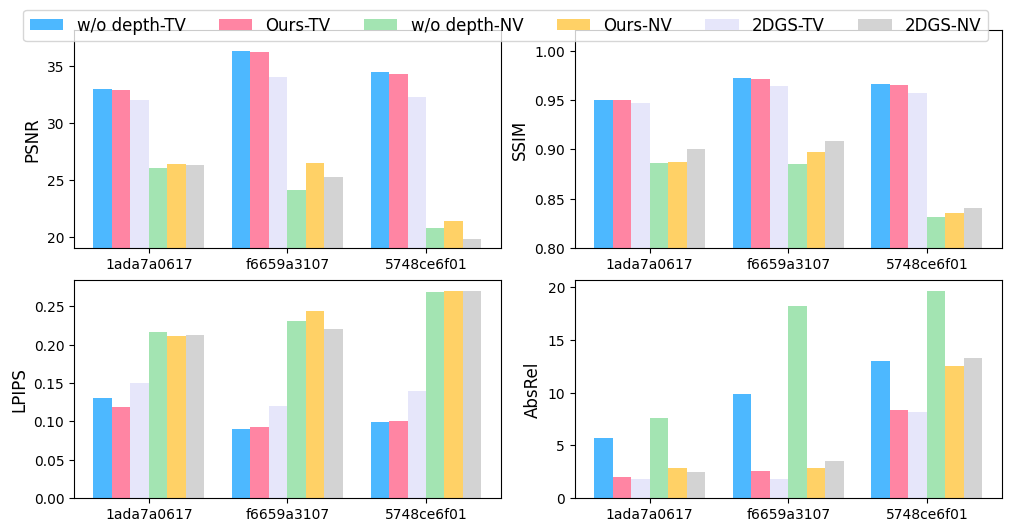}\\
    \caption{\label{Fig_rendering} Rendering quality of RGB images with respect to $L_d$ in Eq.~\ref{eq:rgb_loss} on ScanNet++. "TV" refers to the training view and "NV" to the novel view.}
    \vspace{-0.5cm}
\end{figure}

A high-fidelity novel view synthesis (NVS) capability and a color-geometric consistent model are essential for robotic applications.
We evaluate NVS performance and the alignment effectiveness between appearance and geometry branches. Alignment quality was assessed using the $AbsRel$ metric on the depth maps rendered by the appearance brunch, while NVS performance was evaluated using $PSNR$, $SSIM$, and $LPIPS$ metrics on the NVS test frames provided in ScanNet++. The results are presented in Fig.~\ref{Fig_3dgs}.
Unsurprisingly, the employment of depth improves the model's NVS capability, which is explained by reducing floating Gaussians in free space. However, the additional task constrains the fitting ability of the appearance model, explaining the minor decrease of $LPIPS$ in novel views. We also explore an RGB-D point cloud initialization for 2DGS~\cite{2dgs} representation, which achieves promising NVS results as shown in Fig. \ref{Fig_rendering}.

\end{document}